\documentclass{article}
\usepackage[utf8]{inputenc}

\usepackage{url}
\usepackage{fullpage}

\usepackage{amsmath}
\usepackage{amsfonts}
\usepackage{amssymb}
\usepackage{amsbsy}
\usepackage{amsthm}
\usepackage{latexsym}

\usepackage{bm}
\usepackage{bbm}

\usepackage{complexity}
\usepackage{graphicx}

\usepackage{xcolor}
\usepackage{algorithm}
\usepackage{algorithmic}

\usepackage[round]{natbib}

\newtheorem{theorem}{Theorem}
\newtheorem{remark}{Remark}
\newtheorem{proposition}{Proposition}
\newtheorem{lemma}{Lemma}
\newtheorem{assumption}{Assumption}

\newtheorem{exform}{Example}

\newcommand{\ProofEnd}{\mbox{$\Box$} \vspace{ 8pt}}

\newcommand{\eff}{\texttt{eff}}

\newcommand{\ex}{\mathbb{E}}

\newcommand{\ones}{\mathbbm{1}}
\newcommand{\pr}{\mathbb{P}}

\newcommand\Real{{\mathbb{R}}}
\newcommand\Ints{{\mathbb{Z}}}

\newcommand\bbK{{\mathfrak{K}}}

\newcommand\bbC{{\mathfrak{C}}}
\newcommand\bbV{{\mathfrak{V}}}
\newcommand\cA{{\mathcal{A}}}
\newcommand\cD{{\mathcal{D}}}

\newcommand\cG{{\mathcal{G}}}
\newcommand\cH{{\mathcal{H}}}

\newcommand\cK{{\mathcal{K}}}

\newcommand\cN{{\mathcal{N}}}
\newcommand\ccP{{\mathcal{P}}}

\newcommand\cR{{\mathcal{R}}}

\newcommand\cV{{\mathcal{V}}}
\newcommand\cW{{\mathcal{W}}}
\newcommand\cX{{\mathcal{X}}}
\newcommand\cZ{{\mathcal{Z}}}

\newcommand{\ndctr}[1]{\mathbb{I}{\left\{#1\right\}}}
\newcommand{\tndi}{\rightarrow \infty}
\newcommand{\tndni}{\rightarrow -\infty}
\newcommand{\tndo}{\rightarrow 0}

\def\D{\mathrm{d}}

\newcommand\DefAs{{:=}}

\begin{document}

\title{\bf Distributionally Robust Regularization of Sparse Integer Programming Trained Learning Models}

\author{Sanjeeb Dash, Soumyadip Ghosh, Jo\~ao Gon\c calves, Mark S.\ Squillante\\
IBM Research, Yorktown Heights, NY 10598}

\maketitle

\begin{abstract}
Building explainable machine learning models is crucial for human users to be able to interpret the proposed statistical relationship obtained from the training data. Mixed-integer optimization formulations are often used to train such models with explicit sparsity constraints, aiming to strike the right trade-off between sparsity and prediction accuracy. Existing methods to find the right choice of sparsity -- e.g., via cross-validation -- are computationally expensive. For convex model training formulations, recent advances in distributionally robust optimization (DRO) provide strong generalization while sidestepping this computational burden. We describe an extension of such regularization via DRO to mixed-integer sparse programs, providing statistical guarantees as a function of an associated sparsity parameter of the formulation. We illustrate the use of this approach in the case of building explainable binary classification models using sets of feature value rules.  
\end{abstract}

\section{Introduction}
\label{intro}
%
%
The need for machine learning (ML) models to be easily understandable to humans is critically important in many application domains, whereas many popular ML methods are hard to interpret, analyze, and comprehend.
In applications having societal impact, such as criminal justice, medicine and pharmacology, ML models that are not interpretable may not be satisfactory \citep{rudin19b, rudin22}.
This need has become increasingly critical with the rise of large language models (LLMs), as clearly illustrated by their recent use to settle famous open problems in mathematics; even when formally correct, the resulting arguments are often so impenetrable that even experts in the field cannot follow the reasoning, thus conferring little in the way of understanding~\citep{quanta2026}.
Indeed, twenty-five Fields Medalists have warned of a ``severe misalignment'' in this trend, stressing that a primary goal is conceptual understanding and insight~\citep{fields2026}.
The same lesson applies directly to both ML and LLM models; namely, highly accurate yet humanly uninterpretable outputs leave users unable to trust, verify, or effectively utilize them, underscoring the necessity of explainability and interpretability.

These trends have led in recent years to a considerable emphasis on explainability and interpretability in ML models, including methods such as LIME \citep{ribeiro16} that explain the predictions of a classifier by approximating it locally with an interpretable model.
Such trends have also led to revisiting rulesets \citep{dgw18, rudin17}, rule lists \citep{lakkarajuDecisionSets2016}, and decision trees \citep{rudin19} for interpretable classification, as well as
constructing interpretable ensembles of models hand-picked for explainability~\citep{dsl20}. 

This need for interpretability has resulted in  problem formulations with explicit sparsity constraints in the literature; c.f.~\cite{mazumdar,zhang2023} and~\cite{goh2024}. 
In this paper we consider stochastic optimization problems of the form: 
\begin{subequations}
\label{defn:base_prob}
\begin{align}
L^{\ast}_{\dagger}\; :=\; &\; \min_{v,w} \quad L(v,w) \; =\; \ex_{\xi\sim P^{\dagger}} l(v,w,\xi)   \label{base:obj}\\
&\; \mbox{ s.t.} \quad\;\; g_j(v,w) \; \le \; 0,\qquad j\;\in\;[J] ;\label{base:jt_constr}\\
&\qquad\quad h_m(w) \;\le\; 0,\qquad\;\;\; m\;\in\;[M];\,\label{base:int_constr}\\
&\qquad\quad v\in\Real^d,\; w\in\Ints^{p}. 
\end{align}
\end{subequations}
Here $[J]$ stands for $\{1,\ldots,J\}$.
The function $l(v,w,\xi)$ in the objective~\eqref{base:obj} encodes the loss suffered in representing the data emerging from the true (unknown) distribution $P^\dagger$ using the model parametrized by variables $v$ and $w$. 
The constraints~\eqref{base:jt_constr} are jointly imposed on the real-valued $v$ and the integer-valued $w$, while the constraints~\eqref{base:int_constr} on the integer-valued $w$ may serve the role of enforcing sparsity in the trained model.
Let us now consider a couple of specific examples of our general stochastic optimization formulation~\eqref{defn:base_prob}.

\begin{exform} \label{ex:binclass}
\emph{Binary Classification.} We seek a model 
that correctly classifies a data point $x\;\in\;\cX\;\subseteq \;\{0,1\}^\ell$ as having label $y\in \{0,1\}$. Suppose $\cH =\{h_k(x)\}$ is a collection of functions $h_k\,:\,\cX\rightarrow \{0,1\}$ that each predict the label to be applied to $x$. Construct an ensemble $F(v,x) = \sum_k v_k h_k(x)$ 
and apply the label $\ndctr{F(v,x) > \tau}$ for some pre-selected threshold $\tau\in(0,1)$, where $\ndctr{\cdot}$ is the indicator function that returns 1 if and only if its argument is true. Let $l(v,\xi)$ be a loss function that scores misclassification at data $\xi=(x,y)$ by ensemble $F(v,\cdot)$. 
Further, to impose sparsity on the ensemble, we define $\{0,1\}$-valued variables $w_k$, one for each 
$v_k$. 
The stochastic optimization problem then takes the form:
\begin{subequations}
\label{defn:ensemble}
\begin{align} 
L^{\ast}_{\dagger}\; :=\; & \min_{v,w} \quad \ex_{\xi\sim P^{\dagger}} l(v,\xi) \label{ensemble:obj} \\
 & \; \mbox{ s.t.} \quad \;\; \sum_k v_k \; = \; 1\,, \qquad\; w_k\; \ge\; v_k\;\ge\; 0 \,,\;\;\forall h_k ;\label{ensemble:cvx}\\
 &
 \qquad\quad 
 \sum_k w_k c(h_k) \; \le \; \bbC\, ;\label{ensemble:sparse} \\
 &\qquad\quad 
v_k \in \Real \quad\text{and}\quad w_k \in \{0,1\} \quad \forall \;\; h_k.
\end{align}
\end{subequations}
Constraints~\eqref{ensemble:cvx} ensures the convexity of coefficients $v_k$ and also allow coefficient $v_k$ to be positive only when the corresponding $w_k=1$. Constraints~\eqref{ensemble:sparse} impose a complexity constraint on the ensemble by ensuring that the sum of the individual complexities $c(h_k)$ of each member $h_k$ of the ensemble does not exceed the threshold $\bbC$. 
\hfill \ProofEnd
\end{exform}

Example~\ref{ex:binclass} is studied in greater detail in 
Section~\ref{sec:rules}
where such ensemble models are constructed using disjunctions as the generating class $\cH$ with their specific complexity cost $c(\cdot)$.

\begin{exform} \label{ex:second}
\emph{Sparse Regression Trees.}
Following~\cite{zhang2023}, consider the problem of constructing a regression tree that predicts a real-valued target $y\in\Real$ for a data point $x\;\in\;\cX\;\subseteq\;\{0,1\}^\ell$.
A regression tree is a rooted tree with each leaf labeled by a constant real value and associated with a subset of $\cX$ in such a way that the subsets form a partition of $\cX$; given $x\in \cX$, the predicted value is the label of the leaf associated with the subset of $\cX$ that contains $x$.
Suppose $\cH =\{h_k(x)\}$ is a collection of candidate leaves, where each $h_k\,:\,\cX\rightarrow\{0,1\}$ is a conjunction of feature conditions with $h_k(x)=1$ indicating that $x$ is ``captured'' by leaf $k$, then $v_k$ is the real-valued label of the leaf and $w_k \in \{0,1\}$ is a binary variable such that $w_k = 1$ indicates that leaf $k$ is a part of the final regression tree.
Let $F(v,w,x) = \sum_k w_k v_k h_k(x)$ be the tree predictor.
Let $l(v,w,\xi) = (y - F(v,w,x))^2$ be the squared-error loss that scores prediction error at data $\xi=(x,y)$ by tree $F(v,w,\cdot)$.
Further, a sparsity constraint on the tree can be imposed using the variables $w_k$.
A solution of the following stochastic optimization problem yields an ``optimal'' regression tree of bounded-complexity:
\begin{subequations} \label{defn:regtree}
    \begin{align}
        L^{\ast}_{\dagger}\::=\:& \min_{v,w} \,\ex_{\xi\sim P^{\dagger}}\, l(v,w,\xi) \label{regtree:obj} \\
        & \;\; \mbox{s.t.} \;\; \textstyle\sum_k w_k h_k(x) = 1\,,\;\,\forall x\in\cX\,; \quad\; w_k\,\ge\, \tfrac{1}{B}|v_k|\,\ge\, 0 \,,\;\,\forall h_k ; \label{regtree:cvx}\\
        &\textstyle\sum_k w_k\, c(h_k) \le \bbC\,; \label{regtree:sparse}\\
        & v_k \in \Real \quad\text{and}\quad w_k \in \{0,1\} \,,\; \, \forall h_k.
    \end{align}
\end{subequations}
Constraints~\eqref{regtree:cvx} ensure that the selected leaves partition $\cX$, and thus each data point is captured by exactly one selected leaf, while also allowing the prediction $v_k$ to be nonzero only when the corresponding $w_k=1$ (with $B$ a large constant).
Constraints~\eqref{regtree:sparse} impose a complexity constraint on the tree by ensuring that the sum of the individual complexities $c(h_k)$ of each selected leaf $h_k$ does not exceed the threshold $\bbC$.
Taking the unit cost $c(h_k)=1$ recovers a bound on the number of leaves, which is the complexity measure of~\cite{zhang2023}.
\hfill \ProofEnd
\end{exform}

Example~\ref{ex:second} casts the sparse regression tree of~\cite{zhang2023} in the form~\eqref{defn:base_prob}.
The penalized objective and optional depth restriction used therein are recovered by moving the complexity term~\eqref{regtree:sparse} into the objective~\eqref{regtree:obj} and, respectively, by further constraining the admissible leaves $h_k$.

The only information provided to the user on the true distribution $P^{\dagger}$ of $\xi$  (which models, for example, a data generation process) is a dataset of observations $\cD := \{\xi_n,\,n\in[N]\}$. A natural counterpart to the objective~\eqref{base:obj}
is then to replace the expectation with its empirical average taken over $\cD$. 
One of the foundational concerns of statistical learning is to ensure that the optimal model parameters obtained by solving the empirical counterpart does not overfit the dataset of observations $\cD$.
If only continuous variables were present in~\eqref{defn:base_prob}, 
methods such as LASSO regularization add a weighed regularization term to the objective in order to penalize solutions where many components of $v$ take nonzero values. 
Sparse integer program (IP) formulations~\eqref{defn:base_prob} enforce constraints of type~\eqref{base:int_constr} that may explicitly encode this using the deterministic constraints $h_m$ on $w$. 
While sparsity constraints provide more direct control on regularization, the issue of hyperparameter tuning is not alleviated since, as in the continuous case with the weight of the penalty term, additional parameters (e.g., $\bbC$ in Examples~\ref{ex:binclass}~and~\ref{ex:second}) are introduced that need to be well tuned. Following the standard procedure of cross-validation can be expensive with the need to repeatedly solve the 
IP~\eqref{defn:base_prob}.

As an alternative to address the costs of cross-validation,
techniques from distributionally robust optimization (DRO) have become popular for improved generalization in efficiently selecting models by solving convex programs with continuous-valued variables. There is a rich literature on appropriately parameterizing the DRO formulation of the convex model-selection program to provide statistical guarantees without needing expensive hyperparameter tuning~\citep{bkm19,lz17,nd17}. 
The out-of-sample performance of stochastic optimization formulations has been analyzed~\citep{bkzh19, duchi2018statistics,lam19,lz17} utilizing the empirical likelihood methodology~\citep{owen01}. 
Moreover,~\cite{gsw19} prescribe a general procedure that solves these concave-convex formulations  efficiently to produce models of similar or higher generalization quality on a significantly lower computational budget. 
%
Denote $\ones$ to be the vector of all ones and let $P^0$ be the equal-weight 
distribution $\frac 1 N \;\ones $ over $\cD$. 
The \emph{robust} loss 
$R(v,w)$ and the robust optimal value $R^\ast$ are then given by 
\begin{subequations}
\label{defn:rob}
\begin{align}
R(v,w) \quad&:=\;\; 
	\max_{P\in\mathcal{P}(\rho)} \;\ex_{P} \,l(v,w,\xi), 
  \, \text{ where}
	\,  \ccP(\rho) \: \DefAs \: \left\{ P\,\,|\,\, D(P,P^0) 
 \leq \rho,\; \ones^\top P =1, \;P\ge 0 \right\} ;
\label{defn:rob_loss}\\
R^\ast  \quad&:=\;\; \min_{v,w} \;R(v,w)\,,\quad \text{s.t.} \quad g_j(v,w) \le 0,\;j\in[J];\quad h_m(w)\le 0,\;m\in[M]. \label{defn:rob_prob}
\end{align}    
\end{subequations}

The set $\ccP(\rho)$ is centered at the empirical probability mass function (pmf) $P^0$ and includes all pmfs $P$ that are within a distance $D(P,P^0)$ of $\rho$ from $P^0$, where the distance is measured by
the Kullback-Leibler (KL) divergence:
$D(P,P^0) = -\frac 1 N \sum_n \log (NP_n).$
When $\rho$ is clear from the context, we denote $\ccP(\rho)$ by $\ccP$.
The DRO procedure selects the model that solves for the minimizer $R^{\ast} \::=\:\min_{v,w} R(v,w)$ subject to all the constraints present in~\eqref{defn:base_prob}.
In cases where we only have continuous-valued variables $v$ in~\eqref{defn:base_prob} and $l(\cdot,\xi)$ is convex for fixed $\xi$, it is broadly true~\citep{bkm19,lz17,nd17} that, if $\rho$ is well-chosen, the robust formulation provides a good statistical proxy for the performance under the true distribution $P^{\dagger}$. 
In particular, 
the optimal robust loss value $R^{\ast}$ is an asymptotically (in $N$) valid upper bound on the optimal loss $L^\ast_\dagger$ under the true (unknown) distribution. 
Moreover, we typically need to set $\rho =  {O(\bbV)} / N$, where $\bbV$ is a measure of the explanatory power of the model class. 
Herein we extend this DRO approach in 
important ways in order to effectively apply the DRO approach to solve IP formulations~\eqref{defn:base_prob}.  

\subsection{Our Contributions.}

We present a general framework to regularize the mixed-integer convex program  (MICP)~\eqref{defn:base_prob} based on the DRO counterpart~\eqref{defn:rob}. Theorem~\ref{thm:dromain} in Section~\ref{sec:dro} establishes rigorous  statistical guarantees on how $R^\ast$, the optimal value of the DRO program, is with high probability an upper bound on $L^\ast_\dagger$, the optimal performance under the true (unknown) distribution in~\eqref{defn:base_prob}. This result also provides guidance on how the radius $\rho$ in the DRO formulation should be set in relation with the key sparsity characteristics of the MICP like~\eqref{defn:base_prob} to obtain the desired guarantee, thus extending the DRO literature in this regard to MICP formulations of form~\eqref{defn:base_prob}. 

Of equal importance is the computational challenge that arises in finding good candidate solutions to the DRO formulation of the typically expensive to solve original MICP~\eqref{defn:base_prob}. Unlike DRO formulations of convex programs that have generic meta-heuristic prescriptions~(c.f.~\cite{gsw19}), the DRO-MICP~\eqref{defn:rob} requires case-dependent attention.
%
%
In Section~\ref{sec:rules}, we study the sparse binary classification MICP formulation of Example~\ref{ex:binclass} for the model class of convex ensembles of rulesets. We derive in Theorem~\ref{thm:dro_rules} the implications of our results on the statistical guarantees for DRO formulations of MICPs for this learning problem. Proposition~\eqref{prop:gen_uc_bounds} uses uniform convergence theory from the statistical learning literature to independently derive an upper bound on the generalization performance gap between the true distribution and the empirical distribution over the training dataset. These results together provide important guidance in designing the sparse ensemble heuristic presented in Section~\ref{sec:algo}. The two results also broadly seem to agree with each other in the following sense. The convex program DRO theory also establishes~\citep{lam19,nd17} that, with high probability, an upper bound of the form $ R(v)\, \le\, \ex_{P^0} l(v,\xi) + \sqrt{{\rho \mbox{Var}_{P^0}  l(v,\xi)}}$ holds. While we do not prove this assertion for the MICP~\eqref{defn:base_prob} and its DRO~\eqref{defn:rob}, we note that the $O(\sqrt{\rho})$ dependence of the second term is similar to that observed in Proposition~\ref{prop:gen_uc_bounds} when using the recommended $\rho=O(\bbC)/N$ from Theorem~\ref{thm:dro_rules}.

Section~\ref{sec:algo} constructs a heuristic algorithm to identify good candidate solutions to solve the DRO-MICP~\eqref{opt:ensemble_dro} for the ensemble of rulesets classifier. It produces these solutions by iteratively solving two IPs for variables $(v,w)$ and a concave program over the pmf variable $P$. Finally, in Section~\ref{sec:expts}, we present a small set of results from numerical experiments illustrating that
the sparse convex ensembles of rulesets we construct 
\textit{(a)} match, and in many cases improve, the generalization performance over 
the best state-of-the-art methods; \textit{(b)} are of relatively low complexity leading to easy interpretability; 
and \textit{(c)} are computationally lighter than IP-based methods~\citep{dgw18}.

\textbf{Notation.} 
Let $g() : \Real_+ \rightarrow \Real_+$ be a function.
We say that $g(x) = o(x)$ if $g(x)/x\rightarrow 0$ as $x \rightarrow \infty$. We say that $g(x) = O(f(x))$ if there exists a finite $C>0$ such that $g(x) \le C f(x)$ for all $x > 0$ and that $g(x) = \Theta(f(x))$ if there exists finite $0<C_l < C_h$ such that $C_l f(x) \le g(x) \le C_h f(x)$ for all $x > 0$.

\section{Distributionally Robust Regularization}\label{sec:dro}
%
%

The main result in this section presents an asymptotic statistical coverage guarantee on the optimal performance $L^\ast_\dagger$ of the solution that minimizes the loss objective in~\eqref{defn:base_prob} under the true distribution $P^{\dagger}$.
We obtain this guarantee by solving the robust counterpart~\eqref{defn:rob}, 
where we specifically relate the radius $\rho$ of the probability ball $\ccP$ required to key sparsity characteristics of~\eqref{defn:base_prob}. 

%

We present a brief outline of the approach that will be followed in this section to obtain our main result, namely Theorem~\ref{thm:dromain}.
We will make two important assumptions (formalized shortly). The first is that there are only finitely many values of the integer variables in (\ref{defn:base_prob}) that keep the problem feasible (see Assumption~\ref{asm:int_cfg}) and each such feasible assignment to the integer variables results in a ``nice'' restricted convex optimization problem (\ref{cvx_residual}) (see Assumption~\ref{asm:cvx_rsdl}). 
Statistical coverage guarantees for such convex programs have been established with broad generality~\citep{lz17,bkm19,nd17}. 
In Proposition~\ref{prop:dro_cvx}, we present such a result for the restricted formulation~\eqref{cvx_residual} using solutions of its distributionally robust counterpart~\eqref{cvx_rob}. 
The statistical coverage guarantee for the MICP is then assembled from the individual guarantees for each integral variable restriction using union bounds over the set of all unique configurations of integral variables.

We start with the assumptions that enable our main result. 
\begin{assumption}\label{asm:int_cfg}
The set $\cW = \{w\in\Ints^p\mid h_m(w) \le 0,\,\,m\in[M] \}$ is finite, that is $|\cW|< \infty$. 
\end{assumption}
Assumption~\ref{asm:int_cfg} limits the number of feasible assignments to the integer variables in (\ref{defn:base_prob}) to be finite.
The quantity $\kappa := \log | \cW|$ broadly serves as a measure of the sparsity imposed by the constraints that define $\cW$. 
Lemma~\ref{lem:cW_size_ruleset} in Section~\ref{sec:rules} shows that 
{$\kappa = O\big(\bbC \log (2\ell/\bbC)\big)$}
for the ruleset ensemble binary classification of Example~\ref{ex:binclass}, where
{$\bbC$}
is the desired sparsity factor
{and $\ell$ is the number of binary features.}

%
%
The residual program of~\eqref{defn:base_prob}, after fixing integral variables $w$, is expressed as
\begin{align}\label{cvx_residual}
    L^\ast_{\dagger} (w) \,\,=\,\,\min_{v\in \Real^{d(w)}} \ex_{P^{\dagger}} l(v, \, w,\, \xi) \,,\qquad\text{s.t.} \;\; g_j(v,w) \le 0,\;j\in[J]. 
\end{align}
In typical applications, the sparsity constraints $h_m$ on $w$ are imposed such that any fixed $w$ allows only a small number of the components of the continuous valued $v$ to be active via the constraints $g_j(v,w) \le 0$, with the  
remaining components taking fixed values as imposed by the constraints.
In Example~\ref{ex:binclass}, 
{the choice of $w$ selects at most $\bbC$}
of the coefficients $v_k$ to take nonzero values, with the rest being set to zero. 
Similarly, in Example~\ref{ex:second},
{the choice of $w$ retains at most $\bbC$ of the leaves}
in the problem.
We recognize this sparsity regulation by denoting the effective dimension of $v$ for a fixed $w$ as $d(w)$, so that $v\in \Real^{d(w)}$.
Assumption~\ref{asm:cvx_rsdl} below ensures that an optimal solution to~\eqref{cvx_residual} exists. 
\begin{assumption}\label{asm:cvx_rsdl}
For each $w\in \cW$ and each $\xi$, the functions $g_j(v, w)$ and $l(v,w,\xi)$ are convex in the continuous valued  $v\in\Real^{d(w)}$. Further, the gradients $\nabla_v \ex_\dagger l(v,w,\xi)$ and $\nabla_v g_j(v,w)$ exist and are well defined for each $v$ given a fixed integral $w$, and there exists a point $v_0$ in the strict interior of the constraints, namely it satisfies $g_j(v_0, w) < 0$ for all $j\in[J]$.
\end{assumption}
Under Assumption~\ref{asm:cvx_rsdl},
{the functions $l(\cdot,w,\xi)$ and $g_j(\cdot,w)$ are convex and the strictly interior point $v_0$}
satisfies the Slater KKT conditions.
{Therefore, these KKT conditions are necessary and sufficient for the optimality of the constrained convex program~\eqref{cvx_residual} where a feasible primal solution $v^\ast$ is optimal if and only if there exist multipliers $\lambda^\ast_j \ge 0$}
such that the following equations are satisfied:
%
\begin{subequations}
\label{defn:kkt}
\begin{align}
    \nabla_v \ex_{P} l(v^\ast, w, \xi) \;\;+\;\; \sum_{j=1}^J \lambda^\ast_j \nabla_v g_j(v^\ast,\,w) = 0 ; \label{kkt:stnry}\\
      \qquad g_j(v^\ast, w) \le 0, \; \text{ for } j = 1, \ldots, J ; \label{kkt:feas}\\
{\lambda^\ast_j} \;\;\ge\;\; 0,\;\;\forall j\in [J]\,;\qquad \sum_{j=1}^J \lambda^\ast_j g_j (v^\ast, w) = 0.\label{kkt:cs}
\end{align}
\end{subequations}
Equation~\eqref{kkt:stnry} defines the first-order stationarity condition for the solution $(v^\ast, \lambda^\ast)$ under $P$,
{and represents a system of $d(w)$ scalar equations.}
Condition~\eqref{kkt:feas} imposes feasibility of the primal solution $v^\ast$, whereas~\eqref{kkt:cs} captures the feasibility of the dual $\lambda^\ast$ together with its complimentary slackness property corresponding to the constraints $g_j(v^\ast,w)$. 
We write the KKT conditions~\eqref{defn:kkt} for a general distribution $P$, whereas for the residual program~\eqref{cvx_residual} they are satisfied by the triplet $(v^\ast,\lambda^\ast, P^\dagger)$.
Of course, the true distribution is not available to verify these conditions.
The empirical approximation to~\eqref{cvx_residual} therefore replaces the expectation under the (unknown) true distribution $P^\dagger$ with the empirical mean $\ex_{P^0} l(v, w , \xi) = \frac 1 N \sum_n l(v,w,\xi_n)$. 
An optimal solution $(v^\ast_0, \lambda^\ast_0)$ of this empirical approximation satisfies the KKT conditions~\eqref{defn:kkt} with $P=P^0$.

Define the robust worst case outcome corresponding to the empirical approximation to~\eqref{cvx_residual} as
\begin{align}\label{cvx_rob}
    R^\ast(w) \;\;:=&\;\;\min_{v\in\Real^d}R(v\mid w) \,,\quad\text{s.t.} \quad\;\; g_j(v,w) \le 0,\;\;j\in[J] ,\\
 \text{where}\;\;\;   R (v \mid w) \,\,&:=\,\,\max_{P} \sum_n P_n l(v, \,w,\, \xi_n)\,,\quad\text{s.t.}\;\; -\frac 1 N \sum_n \log \left({NP_n}\right) \leq \rho,\;\; \ones^\top P =1, \;P\ge 0 . \nonumber
\end{align}
The weights $P$ over the collected sample set in the inner maximization that defines the robust loss $R(v\mid w)$ are constrained to lie within a KL-divergence ball of radius $\rho$ about the empirical 
$P^0$. 
For any fixed $v,w$, 
{this inner problem maximizes an objective linear in $P$ over the intersection of the simplex with the sublevel set $\{P : D(P,P^0)\le\rho\}$ of the strictly convex KL-divergence  $D(\cdot,P^0)$.
Suppose the loss $l(v,w,\cdot)$ is not constant over the dataset.
The KL-divergence constraint is then active at optimality, and the strict convexity of $D(\cdot,P^0)$ implies that the maximizer $P^\ast(v,w)$}
is unique (see Proposition~1 of~\cite{gsw19}) and attains
the optimal solution $R(v\mid   w) = \ex_{P^\ast(v,w)} l(v, w,\xi)$.
{The map $v\mapsto R(v\mid w)$ is convex, being a pointwise maximum over $P$ of the functions $\ex_P l(v,w,\xi)$, each convex in $v$ under Assumption~\ref{asm:cvx_rsdl}.
Wherever the subgradient of $l(\cdot,w,\xi)$ reduces to a gradient, the uniqueness of $P^\ast(v,w)$ yields $\nabla_v R(v\mid   w) = \ex_{P^\ast(v,w)} \nabla_v l(v,w,\xi)$ by Danskin's theorem~(Proposition~1 of~\cite{gsw19});
at the remaining points, the same identity holds for subgradients.}
The outer minimization problem~\eqref{cvx_rob} 
{is therefore a constrained convex program, and its optimal solution}
satisfies 
{(the subgradient form of)}
the KKT conditions~\eqref{defn:kkt}
{evaluated at $P^\ast(v,w)$ in place of $P$}.  

DRO theory has been well developed to address statistical guarantees for coverage of the optimal value $L^\ast_\dagger (w)$ of the convex residual program~\eqref{cvx_residual} by the robust optimal $R^\ast(w)$. 
We employ the methodology of empirical likelihood (EL)~\citep{owen01} to derive our results. 
The EL technique evolved as a mechanism to derive convergence and coverage properties for statistical estimators without placing any parametric assumptions on the underlying probability distributions and instead working with pmfs supported on an observed dataset $\cD$. 
The likelihood of observing this dataset under a pmf $P$ over the dataset is $\prod_n P_n$, and this likelihood is maximized over the set of all feasible pmfs by the equal-weight $P^0=\tfrac1N\ones$.
The resulting ratio between the likelihood under any $P$ and that under $P^0$, namely $\prod_n P_n / \prod_n \frac 1 N = \prod_n (NP_n)$, in an important scoring metric. 

A celebrated result in this theory that is extensively employed in DRO theory pertains to convergence (in a logarithmic scale) of the \textit{profile} likelihood ratios of statistical estimators to a $\chi^2$-distribution as $N\tndi$.
The profile likelihood is the maximum likelihood ratio $\prod_n NP_n$ that can be obtained by pmfs $P_n$ over the observed dataset $\cD$ while matching a vector of means under $P_n$ with their true values under $P^\dagger$. 
In the stochastic optimization literature, we naturally consider the case of matching optimization criteria such as the KKT conditions~\eqref{defn:kkt}. 
Let $(v^\ast_\dagger (w), \lambda^\ast_\dagger (w))$ be an optimal solution to~\eqref{cvx_residual} under the true distribution $P^\dagger$ and recall 
$L^\ast_\dagger (w)$ is its optimal value. Define the profile likelihood as
\begin{align}\label{defn:proflk}
\cR (w) \;=\; \cR \left(v^\ast_\dagger (w) , \lambda^\ast_\dagger (w), L^\ast_\dagger(w) \right) \;:= \; \max_P \left\{ \prod_n NP_n\;\;
\left|\quad  \begin{array}{l}
(v^\ast_\dagger (w), \lambda^\ast_\dagger (w), P)\;\,\text{satisfy}~\eqref{defn:kkt},\\
\ex_P l(v^\ast_\dagger (w),w,\xi) \;=\; L^\ast_\dagger (w),\;\;\\
\ones^\top P =1,\,\,P_n \ge 0 .
\end{array}\right.\right\} .
\end{align}
Among the constraints defined in~\eqref{defn:proflk}, there are $(d(w) +1)$ that involve means estimated under pmf 
{$P$, namely the $d(w)$ stationarity equations~\eqref{kkt:stnry} and the single loss value matching equation}.
Define the random vector $X (w) := (\; \nabla_v l(v^\ast_\dagger (w), w, \xi)\,,\; l(v^\ast_\dagger (w) , w, \xi) \,)$, a function of the true distribution $P^\dagger$.
The rank $\tau$ of the covariance matrix of $X(w)$ under the true distribution $P^\dagger$ plays a central role in the next result, which is a restatement of Theorem~3.2 due to~\cite{owen01} (who traces it back to~\cite{ql94}) using our terminology. 
%
\begin{lemma}[Theorem~3.2 of~\cite{owen01}]\label{lem:owen_proflk}
Let $x_1, \ldots , x_N$ be independent random samples of the vector $X(w)$ having finite mean $\mu(w)$ and covariance matrix with rank $\tau \in [1, (d(w) +1) ]$.
Then $-2 \log \cR(w) \Rightarrow \chi^2(\tau)$ as $N\tndi$.    
\end{lemma}

Proposition~\ref{prop:dro_cvx} below provides a statistical coverage guarantee for $L^\ast_\dagger(w)$ by $R^\ast(w)$, 
%
and requires the following assumptions. 
\begin{assumption}\label{asm:cvx_res}
(A) 
{For each fixed $w\in\cW$ and each $\xi$, the loss function $l(v,w,\xi)$ is Lipschitz continuous in $v$, with a Lipschitz constant that applies 
uniformly over $w\in\cW$.}\\
(B) 
{For each fixed $w\in\cW$, 
the constraint functions $g_j(v,w)$ are differentiable in $v$ for all $j\in[J]$.}\\
(C) 
{Under the true distribution $P^\dagger$, the convex residual program~\eqref{cvx_residual} has a unique optimal solution $v^\ast_\dagger(w)$ with associated multipliers $\lambda^\ast_\dagger(w)$ solving the KKT conditions~\eqref{defn:kkt}. Moreover, $\ex_{\dagger} l(v,w,\xi)$ is differentiable in $v$ on a neighborhood $\{|v-v^\ast_\dagger(w)| \le \delta\}$ for $\delta >0$ independent of $w\in\cW$.}\\
(D) The covariance matrix of the vector $X(w)$ has rank $\tau\in [1,(d(w)+1)]$.\\
(E) 
{There exists $\epsilon>0$, which is chosen uniformly over $w\in\cW$ and may be taken to depend on the $\delta$ of~(C), such that for any distribution $P$ with $\sup_{v}|\ex_P l(v,w,\xi) - \ex_{P^\dagger} l(v, w,\xi)|\le \epsilon$ there exists a solution $(v^\ast_P(w), \lambda^\ast_P(w))$ to the KKT conditions~\eqref{defn:kkt} under $P$.}\\
(F) 
{$\frac 1 N\sum_n l(v,w,\xi_n) \rightarrow \ex_{P^\dagger} l(v,w,\xi)$ uniformly over $v$ for each fixed $w$, almost surely.}\\
(G) $\ex_{P^\dagger} [ \sup_v l(v,w,\xi)^2] < \infty$ for each $w\in\cW$. 
\end{assumption}
Assumption~\ref{asm:cvx_res}~(A) 
{does not require each loss function $l(v,w,\xi)$ to be differentiable, only requiring Lipschitz continuity.}
Together with Assumption~\ref{asm:cvx_res}~(C), this allows for the interchange of expectations and gradients when estimating the gradient of the expected loss in the neighborhood of the optimal solution $v^\ast_\dagger(w)$. 
We study Example~\ref{ex:binclass} in detail in Section~\ref{sec:rules} where we use a piecewise convex sample path loss objective, and the algorithm used to find solutions attempts to do so while staying away from the discontinuity in the gradient of the sample path loss function. 
\begin{proposition}
    \label{prop:dro_cvx}
Let the DRO formulation~\eqref{cvx_rob} be defined with radius $\rho = \frac 1 {2N} \chi^2_{d(w) +1, 1-\alpha}$.
Let $R^\ast_N(w) = R^\ast(w)$ be its optimal value, and $L^\ast_\dagger(w)$ be the optimal value of~\eqref{cvx_residual}. 
{Under Assumptions~\ref{asm:cvx_rsdl} and~\ref{asm:cvx_res},}
we have that $\lim\sup_{N\tndi} \pr (L^\ast_\dagger(w) > R^\ast_N(w)) \le \alpha$.
\end{proposition}
The proof of the proposition largely follows the logic of the proof of Theorem~2 of~\cite{lz17}, with the modification that the optimal solution of~\eqref{defn:base_prob} satisfies the KKT conditions~\eqref{defn:kkt}.
We present it here in a concise form and refer the reader to~\cite{lz17} for further details.
\\

\noindent\textbf{Proof of Proposition~\ref{prop:dro_cvx}. }
{First checking that Lemma~\ref{lem:owen_proflk} applies to $\cR(w)$, the estimating equations defining $\cR(w)$ are the KKT system~\eqref{defn:kkt} together with the loss-matching equation, rather than the smooth zero-gradient conditions of the unconstrained setting of
~\cite{lz17}.
By Assumption~\ref{asm:cvx_res}~(A)--(C), the mean of vector $X(w)=(\nabla_v l(v^\ast_\dagger(w),w,\xi),\,l(v^\ast_\dagger(w),w,\xi))$ provides an unbiased estimator of the quantities needed in~\eqref{defn:proflk}. 
Specifically, its first block is bounded by the Lipschitz constant of~(A) and its second has a finite second moment by~(G), and thus $X(w)$ has the desired mean and a covariance matrix of rank $\tau$ by~(D).
Lemma~\ref{lem:owen_proflk} then yields $-2\log\cR(w)\Rightarrow\chi^2_\tau$ as $N\tndi$.}
%

{Consider the event $\cA := \{-2\log \cR(w) \le \chi^2_{(d(w)+1), 1-\alpha}\}$.
Since $\tau \le d(w)+1$, the quantiles satisfy $\chi^2_{\tau, 1-\alpha} \le \chi^2_{(d(w)+1), 1-\alpha}$, and together with the limit $-2\log\cR(w)\Rightarrow\chi^2_\tau$ this yields
\[
\pr(\cA) \;=\; \pr\!\left( -2 \log \cR(w) \le \chi^2_{(d(w)+1), 1-\alpha}\right)
\;\ge\;\pr\!\left(-2\log\cR(w)  \le \chi^2_{\tau, 1-\alpha}\right)\;\overset{N\tndi}{\longrightarrow}\;1-\alpha. 
\]
}
{On the event $\cA$, the maximizer $P_\cR$ of the formulation~\eqref{defn:proflk} that defines the profile likelihood $\cR(w)$ satisfies $D(P_\cR, P^0) = \tfrac{1}{2N}(-2\log\cR(w)) \le \rho$, and thus $P_\cR$ is feasible for the inner maximization~\eqref{cvx_rob} that defines the robust loss $R(v^\ast_\dagger(w) \mid   w)$.}
{The feasibility conditions in the definition of $\cR(w)$ also yield the loss $\ex_{P_\cR} l(v^\ast_\dagger(w), w, \xi) = L^\ast_\dagger(w)$.}
Hence, on $\cA$, we have $L^\ast_\dagger(w) \le R(v^\ast_\dagger(w) \mid w)$.

By construction, the profile likelihood optimal $P_\cR$ also satisfies the KKT conditions~\eqref{defn:kkt} with the true-distribution optimal $(v^\ast_\dagger (w), \lambda^\ast_\dagger(w))$.
Now, restrict the feasible set of weights $P$ in the definition of $R(v^\ast_\dagger(w) \mid w)$ to obtain a formulation:
\begin{align}\label{prop_proof1}
    \max_P \;\;\ex_P l(v^\ast_\dagger(w), w, \xi) \,,\quad\text{s.t.}\quad P\in\ccP(\rho)\;\;\text{and satisfy KKT~\eqref{defn:kkt} at }\; (v^\ast_\dagger(w) , \lambda^\ast_\dagger(w), P).
\end{align}
Again, $P_\cR$ is feasible to this problem, and thus its optimal solution also forms an upper bound on $L^\ast_\dagger(w)$.    
By Assumption~\ref{asm:cvx_res}~(F), the empirical mean loss converges uniformly to the mean loss under the true distribution as $N\tndi$.
{Since $\rho_N = \rho = O(\tfrac1N)$ by assumption, the ball $\ccP(\rho_N)$ shrinks to $\{P^0\}$ as $N\tndi$.}
{The proof of Theorem~2 in~\citep{lz17} yields a stronger uniform bound}
\[
\sup_v \sup_{P\in\ccP(\rho_N)} |\ex_P l(v,w,\xi) - \ex_{P^\dagger} l(v,w,\xi)| \tndo.
\]
{In particular, for all $N$ large enough, every $P\in\ccP(\rho_N)$ satisfies the proximity hypothesis of Assumption~\ref{asm:cvx_res}~(E) with the fixed $\epsilon$ therein.
Therefore the inner minimization $\min_v \ex_P l(v,w,\xi)$ subject to $g_j(v,w)\le 0$, $j\in[J]$, admits a KKT solution $(v^\ast_P(w),\lambda^\ast_P(w))$.}
{The objective of each such inner minimization converges uniformly in $v$ to $\ex_{P^\dagger} l(v,w,\xi)$ by the bound above.
The limit is convex (Assumption~\ref{asm:cvx_rsdl}) with the unique minimizer $v^\ast_\dagger(w)$ (Assumption~\ref{asm:cvx_res}~(C)).
Hence, $v^\ast_P(w)\to v^\ast_\dagger(w)$ as $N\tndi$, uniformly over $P\in\ccP(\rho_N)$.
Then, in the limit, $(v^\ast_\dagger (w), \lambda^\ast_\dagger(w))$ is an optimal solution 
for every feasible $P$ in~\eqref{prop_proof1}.}
In other words, as $N\tndi$, the formulation~\eqref{prop_proof1} is equivalent to
\begin{align}\label{prop_proof2}
    \max_P \;\;\min_v \;\; \ex_P l(v, w, \xi) \,,\quad\text{s.t.}\quad P\in\ccP(\rho)\;\;\text{and satisfy KKT~\eqref{defn:kkt}  at }\; (v^\ast_\dagger(w) , \lambda^\ast_\dagger(w), P).
\end{align}
Thus, 
{on the event $\cA$ and in the limit as $N\tndi$,}
the true optimal value $L^\ast_\dagger(w)$ is upper bounded by the optimal value of~\eqref{prop_proof2}.
{Dropping the constraint that $P$ satisfies the KKT conditions at $(v^\ast_\dagger(w) , \lambda^\ast_\dagger(w), P)$ only enlarges this value, rendering the relaxation $\max_{P\in\ccP(\rho)} \min_{v} \ex_P l(v,w,\xi)$ subject to $g_j(v,w)\le 0$.
The objective is linear in $P$ over the convex compact set $\ccP(\rho)$ and convex in $v$, and thus Sion's minimax theorem yields $\max_P\min_v = \min_v\max_P$ where the latter is exactly $R^\ast(w)$ in~\eqref{cvx_rob}.}
{Therefore $L^\ast_\dagger(w) \le R^\ast(w)$ holds on $\cA$ asymptotically.
Since $\lim_{N\tndi}\pr(\cA) \ge 1-\alpha$, we conclude that $\lim\sup_{N\tndi}\pr(L^\ast_\dagger(w) > R^\ast_N(w)) \le \alpha$, as asserted.} \hfill\ProofEnd

Our main result analyses the optimal quantities
\[
R^\ast \;=\; \min_{w\in\cW} \;R^\ast(w) \qquad\text{and}\qquad L^\ast_\dagger = \min_{w\in\cW} \;\; L^\ast_\dagger (w).
\]
Here, $L^\ast_\dagger$ is the performance under the true distribution $P^\dagger$ of the model that solves~\eqref{defn:base_prob}, and $R^\ast$ is the optimal value of the robust formulation~\eqref{defn:rob_prob}. 
Define $D_{\eff} := \max_{w\in\cW} d(w)$ to be the largest effective dimension among the convex programs that result from restricting $w$ to any feasible $\cW$.
{For Example~\ref{ex:binclass}, the sparsity constraint allows at most $\bbC$ coefficients $v_k$ to be nonzero, and thus $D_\eff = O(\bbC)$.
The embedding dimension $d = |\cH| = \Theta( \;\exp(\bbC\log(2\ell/\bbC)\;)$ is exponentially larger, where $\ell$ is the number of binary features; refer to Lemma~\ref{lem:cW_size_ruleset}.}
\begin{theorem}
\label{thm:dromain}
Choose an $\alpha\in(0,1)$ and let $\tilde{\alpha} = \alpha \, / \,|\cW|\,$. Set 
{$\rho = \chi^2_{ D_{\eff}+1 , 1-\tilde{\alpha}} / (2N)$, where $\chi^2_{ D_{\eff}+1 , 1-\tilde{\alpha}}$ is the $(1-\tilde{\alpha})$-th quantile of the $\chi^2$ distribution with $(D_\eff+1)$ degrees of freedom.}
    Then,  
    we have that 
    \begin{itemize}
        \item[(i)] 
        {$\;\;\rho \le \frac 1 {2N} \left(D_\eff + 1 + 2 \log \left(\frac {|\cW|}{\alpha}\right) + 2 \sqrt{(D_\eff+1) \log\left( \frac {|\cW|}{\alpha}\right)} \right) \;\;$, and}
        \item[(ii)] $\;\; \lim\sup_{N\tndi} \pr\big(L^\ast_{\dagger} > {R}^\ast\big) \le \alpha.$
    \end{itemize}
\end{theorem}


{The proof applies Proposition~\ref{prop:dro_cvx} to each configuration $w\in\cW$ at the reduced level $\tilde\alpha = \alpha/|\cW|$, and then combines the resulting per-configuration guarantees by a union bound over the finite set}
$\cW$ of all possible configurations of integral variables $w$.
{The bound~(i) on $\rho$ follows from the $\chi^2$ tail estimate of Theorem~\ref{thm:ub_chisq},}
quoted here from~\cite{inglot10} who traces this result back to~\cite{laurent00massart}. 
\begin{lemma}[Theorem 2 of \cite{inglot10}] \label{thm:ub_chisq}
{For $Y\sim \chi^2_m$, a $\chi^2$ distribution with $m$ degrees of freedom, and $\zeta\in(0,1)$, let $u(\zeta, m)$ denote the upper-$\zeta$ quantile of $Y$, that is the $(1-\zeta)$-th quantile determined by $\pr( Y \ge u(\zeta, m)) = \zeta$.}
Then, for every $m\ge 1$ and every $\zeta\in (0, 1)$, we have
\[
u(\zeta, m) \le m + 2 \log\left(\frac 1{\zeta}\right) + 2\sqrt{ m \,\,\log\left(\frac 1{\zeta}\right)}.
\]
\end{lemma}


\noindent\textbf{Proof of Theorem~\ref{thm:dromain}.}
The robust optimal of the MICP is given by
{$R^\ast  = \min_{w\in\cW} R^\ast(w)$}.
Let $w^\ast_r = \arg\min_{w\in \cW} R^\ast(w)$ be the related optimal integral variable configuration, 
{so that $R^\ast = R^\ast(w^\ast_r)$}.
Similarly, define the optimal performance under the true distribution as $L^\ast_\dagger \; :=\;\min_{w\in\cW} \,L^\ast_\dagger(w)$.

Define events $E(w) := \{ L^\ast_\dagger (w) \le R^\ast(w)\}$ indexed by each $w\in \cW$. Under the event $\bigcap_{w\in\cW} E(w)$: 
\[
R^\ast \;\;=\;\;R^\ast (w^\ast_r)\;\;\ge\;\; L^\ast_\dagger (w^\ast_r) \;\;\ge\;\; L^\ast_\dagger.  
\]
Thus, the event $\bigcap_{w\in\cW} E(w) \;\subseteq\; \{L^\ast_\dagger \le R^\ast \}  $. Applying complements, we obtain that 
\[
\{L^\ast_\dagger > R^\ast \} \;\;\subseteq\;\; \bigcup_{w\in\cW} E^c(w) \;\;=\;\; \bigcup_{w\in\cW} \{L^\ast_\dagger(w) > R^\ast(w)\}.
\]
{Since $\cW$ is finite (Assumption~\ref{asm:int_cfg}), both the union bound over $w\in\cW$ and the subadditivity of $\lim\sup$ over the finite sum apply, thus yielding}
 \begin{align*}
\lim\sup_{N\tndi} 
\pr \left(L^\ast_\dagger > R^\ast\right) &\;\le\;\; \lim\sup_{N\tndi} \;\pr \left(\bigcup_{w\in\cW} \left\{L^\ast_\dagger (w) > R^\ast(w)\right\}\right)\\
            &\le \sum_{w\in\cW} \lim\sup_{N\tndi} \pr \left(L_\dagger^{\ast} (w) \;\;>\;\; {R}^\ast(w) \right)\\ 
            &=\;\;  |\cW| \;\tilde{\alpha}\;\;=\;\; \alpha. 
      \end{align*}
The last line follows by applying Proposition~\ref{prop:dro_cvx} to each convex program created by fixing the integral variables to $w\in\cW$,
with per-problem level $\tilde\alpha$ and DRO radius $\tfrac 1 {2N} \chi^2_{d(w)+1,1-\tilde\alpha}$ as specified in the theorem statement,
{so that $\lim\sup_N \pr(L^\ast_\dagger(w) > R^\ast(w)) \le \tilde\alpha$ for each $w$.}
{Note that the per-problem radius can be replaced by the common radius $\rho = \tfrac 1 {2N} \chi^2_{D_\eff+1,1-\tilde\alpha}$ specified in the theorem statement, since $D_\eff \ge d(w)$ for each $w\in\cW$ implies $\chi^2_{d(w)+1, 1-\tilde{\alpha}} \le \chi^2_{D_\eff+1, 1-\tilde{\alpha}}$.}
{A larger radius enlarges the ball $\ccP(\rho)$, thus increasing $R^\ast(w)$ and shrinking the event $\{L^\ast_\dagger(w) > R^\ast(w)\}$.
The per-problem coverage bound $\lim\sup_N \pr(L^\ast_\dagger(w) > R^\ast(w)) \le \tilde\alpha$ therefore continues to hold at the common radius, and the careful choice $\tilde{\alpha} = \alpha / |\cW|$ renders the desired result~(ii).}

The upper bound~(i) on the radius $\rho$ of the probability ball is obtained as a straightforward application of Lemma~\ref{thm:ub_chisq}
{with $m = D_\eff+1$ and $\zeta = \tilde{\alpha} = \alpha/|\cW|$, together with $\rho =\tfrac 1 {2N}  \chi^2_{D_\eff+1,1-\tilde\alpha} = \tfrac 1 {2N} u(\tilde\alpha, D_\eff+1)$.  \hfill\ProofEnd}

%
Theorem~\ref{thm:dromain} shows that the optimal solution of the DRO-regularized formulation~\eqref{defn:rob_prob} of the MICP~\eqref{defn:base_prob} asymptotically provides a good upper bound on the best performance achievable under the unknown true distribution. 
This broadly agrees with the DRO theory for convex programs and our result extends it to MICPs of the form studied here. 
The power of Theorem~\ref{thm:dromain} lies in relating the radius $\rho$ needed for this guarantee to the key formulation parameters $D_\eff$ and $\kappa = \log |\cW|$. 
In the next section, we study the binary classification formulation Example~\ref{ex:binclass} in detail.
We show that in this case 
{$D_\eff = O(\bbC)$} and 
{$\kappa = O\big(\bbC\log(2\ell/\bbC)\big)$, where 
$\bbC$ is} the desired sparsity score.
{By bound~(i) of Theorem~\ref{thm:dromain}, the radius of the uncertainty ball then satisfies $\rho = O\big(\tfrac{D_\eff + 1 + 2 \kappa}{N}\big) = O\big(\tfrac{\bbC}{N}\big)$ up to logarithmic terms in $\bbC$. }


\section{Sparse Ruleset Classification}
\label{sec:rules}

Binary classification is the problem of finding a model that correctly classifies a data point $x$ as having label $y\in \{0,1\}$. We assume that data $x$ is as an ordered set of $\ell$ binary-valued features, i.e., $x\in\{0,1\}^\ell$. Data with integral or categorical features can be transformed into this setting by using, for instance, one-hot encoding. Real-valued features can be encoded by binning the range of all observed values. 
%
We consider the model class of convex ensembles of rulesets. 

A \emph{simple rule} takes the form
$r(x) \; \DefAs \; \ndctr{x^j = 0}$ or $ r(x) \; \DefAs \; \ndctr{x^j = 1}$ 
for data $x$ and $j\in [\ell]$. 

A \emph{rule} is a conjunction $t(x)$ of simple rules and has the form
$\; t(x) =  \prod_{v=1}^{C }r_v(x).\;$
The conjunction $t(x)$ checks if a subset of components of $x$ have certain desired values. 

A \emph{ruleset}, or {disjunctive normal form} (DNF) classifier, 
takes the form $\: h(x) \:\DefAs \: t_1(x) \vee t_2(x) \vee \ldots \vee t_M(x)$.
It assigns $x$ a label of $1$ if any rule $t_m$ is satisfied. 
This form allows the classifier to capture nonlinear relationships between the features $x$ and the label $y$. For example, the loan default risk classification problem from the FICO challenge~\citep{fico} applies label `\textit{high risk}' to a disjunction of two conjunction terms 
$\,$
( \#\textit{Loans} $\ge 7$ ) $\vee$ ( (\#\textit{Loans} $\le 5) \:\wedge\: $(\textit{Total Amount} $\ge \$10,000)\,)$.

A \emph{convex ensemble} of rulesets takes the form $F(x) = \sum_{k} v_k h_k(x)$ 
where the nonnegative $v_k$ satisfy $\sum_k v_k =1$. 
The convex ensemble predicts labels by applying a threshold, setting 
$\hat{y} = \ndctr{F(x) \ge 1/2}$.
The quality of the convex ensemble $F$ is determined by the misclassification loss 
at data  $\xi$, and takes values in $[0,1]$ with a piecewise-linear convex form  
\begin{equation}
    \label{def:binclass_loss}
l(v,\xi) \;\;=\;\; \max\,\left\{\;\;0,\,\;\;\; (1-2y) \;\, \left(2 \sum_k v_k h_k(x) - 1\right)\;\;\;\right\}.    
\end{equation}
 
Experimental results in \citep{mug1, mug2} show that, for logic programs (which include rulesets and decision trees), inspection time (the time taken by humans to study and understand the program before applying it) is negatively correlated with human predictive performance applying the logic program. To score the explanatory power of a ruleset, we define its complexity based on the number of rules and the number of terms in each rule. We associate a cost $c(t) = C+1$ with a conjunction $t(x)$ of $C$ simple rules. The cost of a DNF term $h$ is then given by $c(h) = \sum_{m=1}^M c(t_m) = M+ \sum_{m=1}^M C_m$. The ensemble of rulesets similarly bear a cost of the sum of the individual costs $c(F) = \sum_k c(h_k)$. 

Formulation~\eqref{defn:ensemble} in Example~\ref{ex:binclass} provides the complete optimization problem we solve to find a \emph{sparse} convex combination. It introduces $\{0,1\}$-integer variables $w$, one for each disjunction $h_k$ in $F$, and the sparsity constraint is easily expressed using the $w$. The formulation is an extension of the formulation by~\cite{dgw18} who study selection of single sparse rulesets. It uses the objective~\eqref{def:binclass_loss} and introduces binary variables $w_k \in \{0,1\}$ in order to impose a control on the total complexity by imposing constraint~\eqref{ensemble:sparse} on the constructed convex ensemble. Note that the binary $w_k$ are not present in the loss definition. The set $\cW$ of feasible configurations of binary variables $w$ plays a central role in the distributionally robust regulation result in Theorem~\ref{thm:dromain}, so let us first establish its characteristics. 

\begin{lemma}\label{lem:cW_size_ruleset}
Let $\bbC$ be the chosen maximum complexity. Define $\cH := \{h\mid c(h) \le \bbC\}$ to be the set of all disjunctions of complexity up to $\bbC$. Associate a binary variable $w_k$ with each $h_k \in \cH$ and let $\cW \;=\;\{w\mid \sum_{h_k\in\cH} w_k c(h_k) \le \bbC \}$ be the set of all collections of disjunctions $h_k$ such that the total cost is $\bbC$ or less. Then we have that $\log |\cH| = \Theta(\bbC)$ and $\kappa = \log |\cW| = \Theta(\bbC)$.
\end{lemma}

\noindent\textbf{Proof of Lemma~\ref{lem:cW_size_ruleset}}.
For a fixed $\bbC$, there are $\binom{2\ell}{C}$ rules or conjunctions of size up to $\bbC$ that can be used in constructing a disjunctive normal form $h_k$. The complexity cost of each conjunction term is $c+1$. Hence, each $h\in\cH$ is constituted from conjunctions of up to size $\bbC-1$.
%
%
%
%
%
For the number of conjunctions of size $c$, we have the lower bound
\[
\binom{2\ell}{\bbC} = \frac {2\ell}{\bbC}\times \frac{2\ell-1}{\bbC-1} \times \ldots \times \frac {2\ell-\bbC+1}{1} \ge \bigg(\frac {2\ell} {\bbC}\bigg)^\bbC.
\]
We can obtain a corresponding upper bound
\[
\binom{2\ell}{\bbC} = \frac {2\ell}{\bbC}\times \frac{2\ell-1}{\bbC-1} \times \ldots \times \frac {2\ell-\bbC+1}{1} \le \frac {(2\ell)^\bbC} {\bbC!} \le \bigg(\frac {2\ell e} {\bbC}\bigg)^\bbC,
\]
where the last inequality uses the fact that $\exp(c) = \sum_{j=0}^\infty c^j / j!$.
 
Since the number of conjunctions grows exponentially with $\bbC$, there exists a constant $V > 1$ such that 
$ 
|\cH(\bbC) | \le V \cdot \binom {2\ell} {\bbC-1}$. To see this, consider the subset of $\cH$ with one term of size $c<\bbC -1$. Then, this subset is of size at most $\binom{2\ell}{c} \cdot \binom{2\ell}{\bbC-1 - (c+1)} \ll \binom{2\ell}{\bbC-1}$. 
Thus, we also have that 
\begin{equation}\label{bounds:size_H}
(\bbC-1) \log \frac{2\ell} {\bbC-1} \;\le\; \log |\cH| \;\le\; \log V + (\bbC-1) \log \frac {2\ell e} {\bbC-1}.
\end{equation}
The arguments for constructing a single $h\in\cH$ of maximum complexity $\bbC$ from conjunctions of maximum complexity $\bbC$ can be repeated for constructing convex ensembles of complexity $\bbC$ constituted with $h\in\cH$, which gives us the desired result on $\kappa$, that is, $\log |\cW|$. \hfill\ProofEnd    

We next present results from uniform convergence theory in the statistical learning literature to shed light on the expected gap in performance between a ruleset and a convex ensemble of ruleset models under the true data distribution $P^\dagger$ and under an empirical distribution $P^0$ over samples $\cD$ generated from the (unknown) true distribution.
\begin{proposition} \label{prop:gen_uc_bounds}
For the true distribution $P^{\dagger}$ 
and the empirical pmf $P^0$ over a sampled 
set $\cD$: 

(a) For any individual ruleset $h$ with max-complexity $c(h) \le \bbC$, measuring misclassification as $l(h,\xi) = |h(x) - y|$, we have that for any $\delta \in (0,1)$, with probability at least $1-\delta$:
	\begin{align}
		\ex_{P^{\dagger}} [l( h , \xi)] & \;\le\; {\ex_{P^0} [ l( h , \xi)]} 
\;+\; 
  \bigg(\frac 2 N \bigg( (\bbC-1) \log \frac{2\ell} {\bbC-1} + \log \frac{1}{\delta} \bigg)\bigg)^{\frac 1 2} .
    \label{prop_eq:gen_h}
	\end{align}
    
(b) For any convex ruleset combination $F = \sum_k v_k h_k$, define $\bbC'\;\DefAs\; \max_k \{c(h_k)\}$ to be the maximum complexity of its constituent conjunctive terms, and measure loss $l(v,\xi)$ as given by~\eqref{def:binclass_loss}. Then, for any given $\delta\in(0,1)$ and for any $\Delta \in (0,\frac 12)$, we have with probability at least $1-\delta$:
	 	\begin{align}\label{prop_eq:gen_F}
		 P^\dagger (l(v,\xi) \le 0) & \; \le \; P^{0} (l(v,\xi) \le \Delta)
   + 
	O\bigg( \frac 1 {\sqrt{N}} \bigg(\frac {(\bbC' -1)}{\Delta^2}  
 \log N  + \log \frac 1 {\delta}\bigg)^{1/2}\bigg).
   \end{align}
    
\end{proposition}
The proof is presented in Appendix~\ref{apdx:proofs}, and is an extension of the Vapnik-Chervonenkis bounds from~\citet{gblm97} and~\citet{sfbl98}.
The typical model learning approach chooses an optimal model that minimizes the training error, the first term in the upper bound~\eqref{prop_eq:gen_h} or~\eqref{prop_eq:gen_F}. The second term is a ``correction'' factor to account for the true distribution and also forms a proxy for the expected gap in performance when testing the trained model on unseen data. It is independent of the model parameters and grows as $O\left(\sqrt{\tfrac {\bbC} N}\right)$ for an individual ruleset and as $O\left(\sqrt{\tfrac {\bbC'}N}\right)$ for an ensemble model. 
This gap has been observed 
for instance in
the results of~\cite{dgw18}, who pick optimal sparse single ruleset models. 
They show that optimal rulesets satisfying a higher complexity $\bbC$ constraint can reduce the training error to zero but do not generalize as well. 

Proposition~\ref{prop:gen_uc_bounds} shows that convex combinations of rulesets enjoy an important advantage over a single ruleset; the generalization penalty (the second term) grows only with the largest complexity of its constituents, while allowing for extra continuous parameters (the $v_k$) to minimize the training error (the first term). Hence, for a fixed complexity budget, convex combinations of rulesets can significantly improve training performance over single rulesets while displaying the generalization performance of its largest constituent ruleset. This is in agreement with the
extensive literature that has emerged over the past two decades on developing stronger models by taking linear combinations of binary models, which includes ensembling models~\citep{sk95} such as AdaBoost~\citep{fs97} and their variants. An extended treatment is available for collections of ruleset-like decision trees such as random forests~\citep{b01RF}. 

Determining the right model sparsity is important to balance interpretability with improved generalization. 
\citet{dgw18} employ cross validation to pick the best value for $\bbC$, but at the expense of a large computational cost since the procedure involves repeatedly solving a complex combinatorial minimization problem.
We will apply the distributionally robust regularization methodology studied in Section~\ref{sec:dro} as a lower computational cost alternative. The robust optimal formulation corresponding to
the integer program~\eqref{defn:ensemble} is given by
\begin{subequations}
\label{opt:ensemble_rob_fmln}
\begin{align}
R(v,w) \quad&=\quad 
	\;\max_{P}\;\; E_{\xi\sim P} \min\; \{ 0 , \,(1 -2y_\xi) \;(2\sum_{k\in\cH} v_k h_k(x_\xi) - 1)\; \} \label{opt:ensemble_robloss} \\ 
  \quad\text{ s.t. }
	&\;\;  \ones^\top P = 1 , \;\; P\ge 0, \;\; -\frac 1 N \sum_n \log \left(NP_n\right)  
 \le \rho, \nonumber \\
 \text{and} \quad\;\; {R}^\ast\quad &=\quad\; \min_{v,w} R(v,w)\quad \text{s.t.}\;\; e^\top v = 1;\quad \sum_k w_k c(h_k) \;\le\; \bbC \label{opt:ensemble_dro}\\ 
 &\qquad w_k \;\ge\; v_k \;\ge\; 0,\;\;\; w_k \in \;\{0,1\} \forall k\in\cH. \nonumber
\end{align}     
\end{subequations}
For the (fixed) ensemble $F(\cdot)=\sum_{k=1}^{|\cH|} v_k h_k(\cdot)$ over a collection $\cH$,  the robust loss $R(v,w)$ is the worst loss observed for the convex ensemble with weights $v$ over the KL-divergence uncertainty ball $\ccP$ of {\em convex} vectors (i.e., vectors from the unit simplex $\{p \in \Real^N_+ : \sum_{n=1}^N p_i = 1\}$) representing pmfs over the training dataset. 
The radius $\rho$ of the probability ball $\ccP$ is chosen using statistical principles outlined in Theorem~\ref{thm:dromain} that provides guarantees on covering the performance at the true $P^{\dagger}$. In particular, we have the following result.
\begin{theorem}\label{thm:dro_rules}
Choose an $\alpha\in(0,1)$ and let $\tilde{\alpha} = \alpha/|\cW|$. Set $\rho = \chi^2_{\bbC+1,1-\tilde{\alpha}} / 2N$ in the DR regularized formulation~\eqref{opt:ensemble_robloss}. Define its 
optimal value to be $R^\ast$, and the optimal value of the sparse ruleset MICP~\eqref{defn:ensemble} under the true distribution $P^\dagger$ to be $L^\ast_\dagger$. Then, we have that (a) the  
radius $\rho \le \frac {O(\bbC)}  N$, and (b) $\lim\sup_{N\tndi} \pr (L^\ast_\dagger > R^\ast) \le \alpha$.
\end{theorem}


\noindent\textbf{Proof of Theorem~\ref{thm:dro_rules}.}
The sparsity constraint~\eqref{ensemble:sparse} in the binary classification MICP~\eqref{defn:ensemble} only allows at most $\bbC$ integral variables $w_k$ to take value $1$. As a consequence, the effective dimension of the continuous variables $v$ in the restricted convex formulation of form~\eqref{cvx_residual} is bounded by $\bbC$. Thus, $D_\eff = \bbC$. By Lemma~\ref{lem:cW_size_ruleset}, the set of all unique configurations of $w$ satisfies $\log |\cW| = O(\bbC)$. Applying Theorem~\ref{thm:dromain} gives us the desired result. 
\hfill\ProofEnd    


\section{The Sparse Ruleset Ensembling Algorithm}\label{sec:algo}
%
%

We now present our algorithm for identifying the optimal sparse convex DNF combination $F(x)$ that provides a good candidate solution to the DRO formulation~\eqref{opt:ensemble_robloss}. The typical approach to solving minimax formulations is to iterate alternatively between improving the objective with respect to
one of the two sets of variables $(v,w)$ and $P$ while keeping the other fixed. \citet{gsw19} show that this approach works well when all variables in the convex formulation are continuous valued, and that the robust formulation can be efficiently solved to produce models of similar or higher generalization quality on a significantly lower computational budget than cross-validation.
Minimax formulations with integral constraints are known to be hard to solve and form a special case of discrete bilevel optimization. The recent literature on this topic \citep{fm,flms} has focused on solving such formulations with only a moderate number of variables (see also \cite{bls} on bilevel problems involving uncertainty). 

The set $\cH$ of feasible rulesets $h_k$ that obey $c(h_k)\le \bbC$ is of size $O(\exp(\bbC))$ (as shown in Lemma~\ref{lem:cW_size_ruleset}), and thus a straightforward implementation of either the MICP~\eqref{defn:ensemble} or its DRO counterpart~\eqref{opt:ensemble_dro}  over all such rulesets is cumbersome to solve. 
The ensembling literature addresses a similar problem by iteratively picking a new model to add to the candidate ensemble. For instance, the AdaBoost algorithm~\citep{fht00} takes this approach using a selection criterion that can be viewed as minimizing greedily a generalized global loss minimization problem over the set of all possible ensembles. 
(\cite{bkzh19} show that this generalized loss objective for AdaBoost can itself be written as a distributionally robust objective function.)

We follow a strategy of iteratively growing a set of ``good'' DNF models over which the robust ensemble is to be assembled, and extend the column generation approach of~\cite{dgw18} to efficiently explore the large space in picking rulesets.
In each iteration $t$ of our algorithm, we alternate between selecting a new addition to a collection of DNF models $\cH_{t}$ and finding the best sparse ensemble over the collection. The algorithm initializes at iteration $t = 0$ with an empty collection $\cH_{t} = \varnothing$ and data weights $P^{t}$ initialized to $P^0$, the empirical (equal-weights) pmf. 
We next present our heuristic algorithm to find good solutions to~\eqref{opt:ensemble_robloss}.\\
\hspace*{0.2in}\textbf{Sparse Ensembling Algorithm.} \\
\hspace*{0.2in}For $t=1,2,\cdots$:\\
\hspace*{0.4in}$1.\; \; $ Solve for $ \;
        h^{\ast}_t \;=\; \min_h \ex_{P^{(t-1)}}\;l(h,\xi)\,,\quad \text{s.t.}\;\;\; c(h) \le \bbC' \; $ \\
\hspace*{0.4in}$2.\; \; $ Update $\;
        \cH_t = \cH_{t-1} \cup \{h^{\ast}_t \}$ \\
\hspace*{0.4in}$3.\; \; $ If $ |\cH_t| \le 5$, skip to Step 5. Else, solve for best sparse ensemble:\\
\hspace*{0.6in}For $m=2,3,\ldots,6:$\\ 
\hspace*{0.8in}$i:$ 
Solve for $ \; L^\ast_m \;$, the optimal value of~\eqref{opt:sparse_subprob} 
over set $\cH_t$ with budget $m\bbC'$\\
\hspace*{0.8in}$ii:$ Stop with respect to $m$ if objective $L^\ast_m > (1.-0.005)\times ($previous best$)$\\
\hspace*{0.6in} Set $m_t\leftarrow m$ and let $\{w^\ast_{t,k}\}$ be the optimal sparse weights \\
\hspace*{0.6in} Set ensemble $F_t \leftarrow \sum_{k=1}^{|\cH_t|} w^\ast_{t,k}\,\, h^\ast_k$ whose training loss is $L^\ast_{m_t}$ from 3$i$\\ 
\hspace*{0.4in}$4.\; \; $ If $\;\;L^\ast_{m_t} < (1.-0.005)\times L^\ast$, then: \quad
Set $L^\ast \leftarrow L^{\ast}_{m_\ast}$ and $t^\ast \leftarrow t$\\
\hspace*{0.4in}$\quad\;\; $ Else $\;\qquad\;$ if $(t - t^\ast) > 20,\;\;$ then:\quad
Stop with respect to $t$ and \textbf{return} $L^\ast$ and $t$ \\
\hspace*{0.4in}$5.\; \; $ Set $ \; F'_{t}(x) = \frac{1}{|\cH_t|} \sum_{k=1}^{|\cH_t|} h^{\ast}_k(x)\,, \;$ \;\;\; and \;\;\;
solve for $ \; P^{t}  = \arg\max_{P\in\ccP} \;\ex_P\, l(F'_{t}, \xi) $
In Step 1, a modified version of the column generation approach studied by~\cite{dgw18} is used to iteratively construct a ruleset of complexity up to $\bbC'$ that minimizes loss under our chosen data weights $P^{t-1}$. This step, outlined in Section~\ref{ssec:cg}, is computationally the most expensive because of the structured exploration of the $O(\exp(\bbC'))$-sized space of 
rulesets. A smaller $\bbC'$ complexity ruleset is desired in this step as advised by Proposition~\ref{prop:gen_uc_bounds}(b), which shows that test performance of a convex ensemble is dependent on the largest complexity of the constituent rulesets. 

Step 3 solves for a sparse ensemble over the collection $\cH_{t}$ with the IP formulation~\eqref{opt:sparse_subprob} as described in Section~\ref{ssec:sparse_ens}. It iteratively goes through the best  ensemble constructed from $m=2,\ldots,6$ disjunctions out of $\cH_t$. It stops when the quality of the solution does not improve more than 0.5\% percent. The selected ensemble has complexity $m_t\bbC$ and assigns weights $w^\ast_{t,k}$, and its loss performance is $L^\ast_t$.

%
%

Step 4 ensures that the outer iterations $t$ progress only if sufficient progress is being made in constructing sparse ensembles with improving performance $L^\ast_t$ on training data.
Step 5 interleaves with the inner maximization in the robust objective~\eqref{opt:ensemble_robloss} that (as provided in Section~\ref{ssec:dro}) optimizes for the worst pmf $P^t$ assigned to the 
training dataset given an equal weight ensemble 
of rulesets in $\cH_t$. Using a dense ensemble in this step yields the best results in our numerical experiments discussed in Section~\ref{sec:expts}. We outline the subproblems that are solved in Steps 1, 3 and 5, respectively.

\subsection{Column Generation for Step 1.} \label{ssec:cg}
This formulation searches over the (exponentially large) space of rulesets of complexity up to $\bbC'$. We extend the approach of~\citet{dgw18} to efficiently explore this space using a column generation reformulation to handle 
the general case where the data weights are any arbitrary pmf $P^{t-1}$. Let $\cK$ be the set of all possible rules that can be formed using 
the $\ell$ features of $x$. 
For each $k \in \cK$, let $Z_k$ be the set of indices of data points with label 0. For each data index $n$, let $\cK_n$ be the set of indices $k$ such that \emph{rule} $k$ is satisfied by data point $x_n$. Let $\cD^1$ be the subset of indices $[N]$ of the training dataset $\cD$ that represent data labeled with $y=1$. Further let $c_k$ be the complexity of rule $k$.
We then solve the formulation
\begin{subequations}
    \label{opt:cg}
\begin{align}
    \min \quad & \sum_{n \in \cD^1} \xi_n + \sum_{k \in \cK} \sum_{n \in Z_k} P_n w_k \label{master-obj}\\
    \mbox{s.t.} \quad & \xi_n + P_n \sum_{k \in \cK_n} w_k \geq P_n, \;\; \xi_n \geq 0, \;\; n \in \cD^1; \label{master-const1}
    \\
    & 
    \;\;\;\; \sum_{k \in \cK} c_k w_k \leq C, 
    \;\; 
     w_k \in \{0,1\}, \;\; k \in \cK. \label{master-bin}
\end{align}
\end{subequations}

This formulation chooses a set of rules (given by nonzero values of $w_k$) that does not minimize the $0$-$1$ loss but instead minimizes a weighted Hamming loss: for each data point $x_n$ with label $1$, $\xi_n$ provides the exact loss weighted by $P_n$, while the second objective term with $P_n$ values set to $1$ provides the number of rules in the chosen ruleset that yield incorrect values for $0$-label data points.  Since the number of rules is exponentially large, we solve this formulation via column generation as in \citep{dgw18} except that the terms in the dual constraints for the linear program (LP) relaxation of~\eqref{opt:cg}
are scaled by the values $P_n$.
Appendix~\ref{apdx:cg-details} 
provides further details on our column generation method. 

\subsection{Selecting Sparse Convex Ensembles in Step 3.}\label{ssec:sparse_ens} 
We select an optimal
ensemble from within the collection $\cH_{t}$ of DNF models.
%
Let $\cD^0$ and $\cD^1$ be the subset of indices $[N]$ of the training dataset $\cD$ that represent data labeled with $y=0$ and $y=1$, respectively. 
For each data index $n$, define $\cK_n \,:=\,\{ k\,|\, h_k(x_n) = 1,\,\,k\in[t] \}$ to represent the set of all DNF {rulesets} that predict a label of $1$ at $x_n$. Auxiliary variables $\eta_n$ for each data index $n$ define the piecewise misclassification loss $l(v,\xi_n)$. We then have 
the following IP:
\begin{subequations}
\label{opt:sparse_subprob}
\begin{align}
    \min_{v,\eta} &\;\;\sum_{n\in [N]} p_n \eta_n 
   \quad \text{where}\;\; 
   \quad  \ones^\top v = 1; 
   \;\;\;\; \eta_n \ge 0 , \;\;\forall n\in[N]; \; \quad w_k \in \{0, 1\} , \;\;\forall k\in[t] ; \label{opt:sparse_sp_obj}\\
\text{s.t.} 
&\quad
\sum_{k\in\cK_n} v_k \,\,- \left(\frac 1 2 - \Delta \right) \,\le\,  \eta_n,\;\forall n \in \cD^0 ;\quad \left(\frac 1 2 + \Delta\right) - \sum_{k\in\cK_n} v_k \,\le\,\eta_n ,\;\forall n \in \cD^1 \label{opt:ens-slacks} \\   
&  \quad\quad v_k \le w_k , 
    \;\; \sum_k w_k  c(h_k) \;\le\; \bbC.     \label{opt:ens-intvar}
\end{align}
\end{subequations}
The constraints~\eqref{opt:ens-slacks} define the piecewise-linear loss function $l(v,\xi_n)$; 
the form~\eqref{def:binclass_loss} was chosen to ensure that this problem is reformulated as a linear IP, a form that is handled stably by commercial solvers. 
The form in~\eqref{def:binclass_loss} is  generalized here by incorporating an extra parameter $\Delta$ to strictly separate the two data subsets, and this is in line with the ensemble generalization result in Proposition~\ref{prop:gen_uc_bounds}(b). Our experiments in Section~\ref{sec:expts} show that a positive $\Delta$ is necessary and that a small separating value, such as $\Delta = \bbC / (2 \cdot\min_k c(h_k))$, suffices.  Binary variables $w_k$ introduced in~\eqref{opt:ens-intvar} enforce the complexity bound on the chosen convex combination. 
The IP formulation~\eqref{opt:sparse_subprob} is relatively small 
with only as many binary variables as the number of DNF models $h_k$ in the current collection.

\subsection{Solving for Worst-Case PMF in Step 5.}\label{ssec:dro} 
The objective function of~\eqref{opt:ensemble_robloss} is given by $\sum_{n\in[N]} l(v^t,\xi_n)\cdot P_n$,
where the $v^t$ are the weights assigned to the DNFs in $\cH_t$, and the individual loss values $z_n = l(v^t,\xi_n)$ are constructed as given by~\eqref{def:binclass_loss}. We assign equal weights $v^t = (1/t)$ to the members of the DNF collection $\cH_t$ and so $z_n \in \{-1,-1+2/t,\ldots,1\}$. We choose not to use the sparse ensemble generated in Step 3 because our numerical results showed that their use can lead to the scenario where the worst-case pmf iterates $P^{t}$ are not sufficiently distinct to produce in turn enough distinct DNF models in Step 1 and the iterations stall. This is partly the result of the (piecewise) linear form~\eqref{def:binclass_loss} for the loss in order to yield a linear IP form in the formulation~\eqref{opt:sparse_subprob}. 

The concavity of the $\log$ function ensures that the feasible set $\ccP$ is strictly convex, and thus  formulation~\eqref{opt:ensemble_dro} is a concave program with a linear objective.
An effective algorithm to efficiently identify a solution to this concave formulation has been studied in broad generality for $\phi$-divergences by~\cite{gsw19}. Their Procedure 1 in Section A.2 is developed based on Lagrangian optimization theory~\citep{lbgr69}, which we employ in our experiments below. 
Importantly, Proposition 2 of~\cite{gsw19} also establishes a worst-case bound on the computational effort of $O\left(N \log N  + \log^2\left(\frac 1 \epsilon\right)\right)$ required to obtain an $\epsilon$-optimal solution to~\eqref{opt:ensemble_dro},
where $\epsilon$ is a small machine-precision parameter that does not relate to any formulation or algorithm parameters.

\section{Numerical Experiments}\label{sec:expts}

Following the numerical experiments of~\cite{dgw18}, we conduct numerous tests on
classification datasets from the UCI repository~\citep{ucirepo} and the dataset from the FICO Explainable Machine Learning Challenge~\citep{fico}, which uses individual credit history to predict loan repayment risk. Appendix~\ref{apdx:expts} provides the
details on the datasets and how they were preprocessed, including any transformation of non-binary features.

\textbf{Competing Methods.} We compare the test performance and model complexity of our DR regularized ruleset induction approach (DR) with the sparse ruleset algorithm (CG) of~\citet{dgw18}. We also include results for three decision tree models: the CART~\citep{Breiman1984ClassificationAR} and random forest~\citep{b01RF} (RF) methods presented in~\citep{dgw18}, and the RuleFit method of~\cite{friedman2008}. The RF method often produces models with good generalization. But it does so using dense ensembles of decision trees that are not appropriate from an explainability viewpoint, hence its model complexity results are omitted. While the RuleFit method does allow the user to suggest a desired complexity target, the actual complexity of the produced model can be quite large (see the "Interpretability" discussion below) and hence its complexity is also omitted.

\textbf{Implementation Details.}
We implement the sparse ensembling heuristic in Java, using version $22.1$ of the commercial solver IBM ILOG CPLEX to solve the LPs and MILPs.
Our column generation implementation is an extension of the approach
in~\citep{dgw18}; refer to Section~\ref{ssec:cg} and 
Appendix~\ref{apdx:cg-details} for more details. 
The CG algorithm converges to the optimal of the LP relaxation of the IP~\eqref{opt:cg} when given enough time. 
However, to maintain a reasonable overall running time, we limit the time spent 
on the pricing problem in each iteration to $30$\textit{s}, the time spent on the overall column generation process to $300$\textit{s}, 
and 
the number of column generation iterations to five. 

Appendix~\ref{ssec:apdx-expt-dr} describes tuning experiments which showed that setting a low value for $\bbC'=5$ grants more generalization flexibility to the sparse ensembling formulation~\eqref{opt:sparse_subprob}. 

Theorem~\ref{thm:dromain} recommends that the KL-divergence radius be set as $\rho = \tfrac 1 {2N} \chi^2_{{D_\eff+1}, 1-\tilde{\alpha}} = O\left(\frac {\bbC}N\right)$. 
For the test datasets, this mean a $\rho$ in  $O(10^{-2})$ to $O(10^{-1})$ 
, and we set $\rho=0.05$ in all experiments. 

Each input dataset is split uniformly at random into two datasets~--~the training subset that contains $90\%$ of the data 
and the testing subset that contains the other 
$10\%$. 
A total of $10$ such permutations are considered to produce the reported means and confidence intervals (CIs).

\noindent\textbf{Interpretability.}
As a representative example, we compare the output models of CG, DR and RuleFit over the same data partition on the ILPD dataset to illustrate their differences with respect to human interpretability. 
The CG method provides that all patients whose data satisfy the rule\\
\hspace*{0.8in}``(D Bilirubin $\le 0.2)\, \wedge \,(25 \le $ Alamine A $\le 35)\,$ $ \wedge $ (Albumin $>3.3$)'' \\
are at high risk of liver cirrhosis. The DR algorithm improves on this classification with a `scorecard' of weighted rulesets. Specifically, it assigns weights to rulesets as:\\
\hspace*{0.1in} $0.62$ to the ruleset identified by the CG method, and\\
\hspace*{0.1in} $0.38$ to ``(D Bilirubin $\le 0.6)\,\wedge\,($Aspartate A $\le 28.0)$ $ \,\wedge$ (Tot Proteins $\le 6.6) \,\wedge\,($A/G Ratio $\le 0.9)$''\\
The user applies this classifier by summing up the total weight assigned to the patient's data, where each ruleset provides its assigned weight if the input matches the rule. The patient is classified to be at high risk if the total weight exceeds the threshold of $0.5$. Each ruleset in this example happens to be chosen as a conjunction of four inequalities, and thus has a cost of $5$ (adding $1$ for each ruleset count). Hence, the costs of the CG model is $5$, and the DR scorecard is $10$. 

The RuleFit method constructs a relatively more complex regression model using all (normalized) features and up to three rulesets as the explanatory variables. LASSO sparsification 
obtains a model with nonzero intercept and coefficients for nine of ten features. In addition, it assigns coefficients\\
\hspace*{0.4in}$+0.65\;\;$ to $\;\;$``(D Bilirubin $\le 1.0)\,\wedge\,($A Phosphotase $\le 208.0)$'',\\
\hspace*{0.4in}$+0.17\;\;$ to $\;\;$``(D Bilirubin $\le 1.0)\,\wedge\,($A Phosphotase $> 208.0)$'', and \\
\hspace*{0.4in}$-0.73\;\;$ to $\;\;$``(D Bilirubin $> 1.0)$''.\\
The linear combination of the intercept and individual feature values (scaled by a coefficient) plus the combinations of the (0/1) rule values (scaled by a coefficient) are applied to patient data. A patient is classified as at high risk if the total exceeds $0.0$. Even assigning a cost of $1$ for each nonzero linear term and adding up the complexity of the rules used results in a high complexity for RuleFit of $18$.
We believe that even this high number does not properly capture the difficulty of interpreting RuleFit rules. It seems hard to interpret a weighted sum of two different real-valued features, e.g., 1.65 D Bilirubin + 0.92 Phospotase.

\begin{table*}[!tbhp]
\caption{(Left) Classification Performance over Unseen Test Data by Five Competing Methods and (Right) Complexity of Models from Three Explainable Methods for Seven UCI Repository Datasets and the FICO Dataset. 
(Last Row:) Mean Rank of Competing Methods (exclude heart).
All Results presented with Mean and $95\%$ CI. 
}
\begin{center}
\resizebox{\textwidth}{!}{%
\begin{tabular}{|l||c|c|c|c|c||c|c|c||}
\hline
Dataset &\multicolumn{5}{c||}{Test Performance (\%)}&\multicolumn{3}{c||}{Model Complexity}\\
Name &DR&CG&CART&RF&RuleFit & DR & CG & CART\\
\hline
heart 
& 81.0(4.4) & 78.9(4.7) & \textit{\textbf{81.6}}(4.7) & \underline{82.5}(1.4) & 
& 22.0(4.2) & \textbf{\underline{11.3}}(3.5) &  32.0(15.9) 
\\
ILPD 
& \textbf{\underline{71.7}}(2.4) & \textit{69.6}(2.4) & 67.4(3.1) & \textit{69.8}(1.0) & 69.1(1.5)
& \textbf{\underline{10.5}}(0.9) & \textit{10.9}(5.3) & 56.5(21.4)
\\
FICO 
& {71.8}(0.6) & 71.7(1.0) & 70.9(0.6) & \underline{73.1}(0.2) & \textbf{72.7}(0.6) 
& 21.5(3.9) & \textbf{\underline{13.3}}(8.0) & 155.0(53.9) 
\\
ionosphere
& \textit{\textbf{93.2}}(3.5) & 90.0(3.5) & 87.2(3.5) & \underline{93.6}(1.4) & 89.2(3.7) 
& 26.5(3.4) & \textbf{\underline{12.3}}(5.9) & 46.1(8.2)
\\
liver 
& \textbf{\underline{62.0}}(2.8) & \textit{59.7}(4.7) & 55.9(2.7) & \textit{60.0}(1.6) & \textit{61.5}(5.2)
& 29.0(1.9) & \textbf{\underline{5.2}}(2.4) & 60.2(30.6) 
\\
pima & \textbf{\underline{76.7}}(2.8) & \textit{74.1}(3.7) & 72.1(2.5) & \textit{76.1}(1.6) & \textit{75.8}(3.4) 
& 22.0(4.2) & \textbf{\underline{4.5}}(2.5) & 34.7(11.4)
\\
transfusion
& \textbf{\underline{79.0}}(2.1) & \textit{77.9}(2.7) & \textit{78.7}(2.2) & \textit{77.3}(0.6) & \textit{77.5}(1.2)
& 13.0(1.5) & \textbf{\underline{5.6}}(2.4) & 14.3(4.5)
\\
WDBC
& 94.9(1.6) & 94.0(2.4) & 93.3(1.8) & \textbf{\underline{97.2}}(0.4) & \textbf{\underline{97.2}}(1.2) 
& 29.0(1.9) & \textbf{\underline{13.9}}(4.7) & \textit{15.6}(4.3) 
\\
\hline
Mean Rank 
& \textbf{1.714} & 3.571 & 4.571 & 2.143 & 2.857 
& 2.000 & \textbf{1.143} & 2.857 
\\
\hline
\hline
\end{tabular}
}
\end{center}
\label{tab:perf-comp}
\end{table*}

\textbf{Test Performance.} Table~\ref{tab:perf-comp} presents the test accuracy and the complexity of the models identified by the various methods
(RuleFit results for heart are omitted because the implementation provided by~\cite{RuleFitCode} does not support categorical data in the dataset used).
For both sets of results, the method with the best average outcomes among the interpretable models is highlighted in \textbf{bold} and the best average outcome overall is \underline{underlined}. Any method with average outcomes within the CI
of the overall best outcome is highlighted in \textit{italics}.
We also report the mean rank of each method,
following
an established approach for statistical comparisons of multiple classification algorithms over multiple 
datasets~\citep{demsar2006}. Given a metric, a method has rank 1 for a dataset if it yields the best value of the metric, rank 2 if it yields the second best value of the metric, and so on. The mean rank is simply the mean of the rank values across different datasets.

The results in Table~\ref{tab:perf-comp} (left) show that DR provides the best overall generalization performance among all methods in half of the eight cases, with the non-interpretable RF method performing best in the remaining four cases.
Moreover, the solution quality of DR is: \textit{(a)} statistically identical to RF, as well as best among the competing explainable methods,
in one of the four remaining cases (ionosphere); and \textit{(b)} statistically identical to the best explainable method for the other three remaining cases (heart, FICO, WDBC).
DR also provides the best mean rank among all methods across all datasets, further reflecting the best method or statistically identical to the best method, with RF providing the second best mean rank followed by RuleFit, CG and CART.

We also compare the complexity of the models from our DR approach with those obtained by three other methods that produce interpretable models. 
The results in Table~\ref{tab:perf-comp} (right) show that the complexity of the estimated models from DR is within a reasonable range of the CG method, even providing the best complexity in one case (ILPD),
while consistently better than CART (except for WDBC) and often significantly so.
The CG method provides the best mean rank among all methods across all datasets, with DR 
providing the second best mean rank followed by CART.
These complexity results illustrate the efficacy predicted by Theorem~\ref{thm:dro_rules} and also by Proposition~\ref{prop:gen_uc_bounds}(b), which showed that sparse convex mixtures of $\bbC'=5$ rulesets have greater flexibility in reducing training error and simultaneously maintain a low generalization gap. 
In contrast, the individual rulesets produced by CG use higher complexity than $5$ to obtain a reasonable balance between reduced training error and lower test efficacy, but overall do not perform as well on test data as the convex combinations generated by DR. While CG may improve its training power with a higher $\bbC$ bound, it may suffer a higher generalization gap as shown in Proposition~\ref{prop:gen_uc_bounds}(a). Hence, it is not clear that the gap in test performance can be closed for any value of $\bbC$ by the CG method. 

Solving the IP formulations~\eqref{opt:cg} are the most computationally expensive steps in the CG and DR methods. 
The sparse convex-combination IP formulation~\eqref{opt:sparse_subprob} of the DR method is typically solved in milliseconds. While CG solves~\eqref{opt:cg} repeatedly in a cross-validation procedure to  fine tune the max-complexity parameter, DR solves a sequence of~\eqref{opt:cg} in 
the iterations of our ensembling algorithm
to obtain the best sparse ensemble. In our experiments, CG ran a factor of $4$ to $6$ times
longer to find its output model (details available in Table~\ref{tab:compute} of Appendix~\ref{apdx:expts}). CART, RF and RuleFit also produce ensembles of rulesets, and use computationally moderate heuristics to generate the included rulesets, but struggle to provide direct trade-offs between sparsity and performance as provided in the core formulation~\eqref{opt:cg} of the CG and DR methods. 


The appendices provide additional results that vary the parameters in this section.
A larger complexity cutoff of $60$ is considered and we see similar generalization performance but, as can be expected, 
at the expense of more complex (and harder to interpret) models. Results for a larger DRO parameter $\rho=0.5$ are presented, illustrating that this parameter provides 
additional benefits in some cases.
Stopping criteria parameters are varied showing
that, with a higher computational budget, slightly higher quality solutions can be obtained by DR for the same complexity cutoff of $30$.
DR iterate trajectories are also provided.

\newpage

\appendix

\section{Theoretical Analysis: Generalization of Rulesets and their Ensembles}\label{apdx:proofs}
The DNF models $h\in\cH$ produce binary output $h \,:\; x \rightarrow y\in \{0,1\}$, where $\cH$ denotes the space of all DNF rulesets of complexity $\bbC$ or smaller. The data $(x,y)$ follow the true (unknown) distribution $P^{\dagger}$, and we are given only an $N$-sized sample dataset $\cD=\{(x_n, y_n),\,\,n\in [N]\}$ generated from $P^{\dagger}$, using the notation $[N] := \{1,\ldots,N\}$. The empirical equal-weight probability mass function
on $\cD$ is denoted by $P^0$.  
In the sequel, we use the shorthand $\ex_\dagger$ and $\pr_\dagger$  for $\ex_{P^{\dagger}}$ and $\pr_{P^{\dagger}}$, respectively, and similarly $\ex_0$ and $\pr_0$ for $\ex_{P^0}$ and $\pr_{P^0}$.


We now establish the following proposition from the main body of the paper. We restate the Proposition in two parts.

\textbf{Proposition~\ref{prop:gen_uc_bounds}(a).}
\textit{
Let $P^{\dagger}$ be the underlying true distribution and $P^0$ the empirical
probability mass function
over a training set $\cD$ of size $N$, with the loss measured as $l(h,\xi) = |h(x) - y|$. 
Then, for any $\delta \in (0,1)$, there exists constants such that for all rulesets $h$ with max-complexity $c(h) \le \bbC$, we have with probability at least $1-\delta$:
	\begin{align}
		\ex_{{\dagger}} [l( h , \xi)] & \;\le\; {\ex_{0} [ l( h , \xi)]} 
\;+\; 
  \bigg(\frac 2 N \bigg( (\bbC-1) \log \frac{2\ell} {\bbC-1} + \log \frac{1}{\delta} \bigg)\bigg)^{\frac 1 2} .
    \tag{\ref{prop_eq:gen_h}}
	\end{align}
}

\noindent\textbf{Proof of Proposition~\ref{prop:gen_uc_bounds}(a).}
Since both $h$ and $y$ are binary, the loss function takes binary values and indeed satisfies $l(h, \xi) = |h(x)-y| = \ndctr{h(x) \neq y}$. For the true (unknown) data distribution $(x,y)\sim P^{\dagger}$ over the space of features and labels, let $p_h = P^{\dagger} (h(x)\neq y)$ be the probability that the loss is $1$ under the true distribution. Define the random variable $Z_h := \ndctr{h(x) \neq y}$ as a Bernoulli random variable with the probability of success (misclassification) $p_h$. 
The training dataset $\cD$ can also be viewed as a set of $N$ independent draws of Bernoulli random variables $Z_n$, 
and $\hat{p}_{h,N} = \frac 1 N \sum_{n\in [N]} \ndctr{ Z_n = 1 }$ is an $N$-sample approximation of the probability $p_h$. 

For each $h\in\cH$, the right-tail Chernoff bound for Bernoulli random variables yields
\[
\pr(p_h -\hat{p}_{h,N} \ge \lambda ) \le \exp(-N \lambda^2/ 2),
\] 
where the probability is taken with respect to the sampling of $\cD$ from $P^{\dagger}$. To obtain a uniform bound that applies to \emph{all} $h\in \cH$, we employ the union bound principle that $\pr (\cup_i A_i) \le \sum_i \pr(A_i)$ for events $A_i$. Thus, we have
\[
\pr\left( \; \exists h\in \cH \;\;\;|\;\;\; p_h - \hat{p}_{h,N} \ge \lambda \,\right) \le |\cH| \exp(-N \lambda^2 / 2).
\] 
Using the lower bound in~\eqref{bounds:size_H}, we obtain the desired generalization result when $\lambda$ is set as
\[
\lambda = \bigg(\frac 2 N \bigg( (\bbC-1) \log \frac{2\ell} {\bbC-1} + \log \frac{1}{\delta} \bigg)\bigg)^{\frac 1 2}.
 \]
 \hfill\ProofEnd

%
%
We next consider convex combination models of the form $F(x) = \sum_{h\in\cH} v_h h(x)$, where the real-valued $v_h$ are nonnegative and $\sum_h v_h = 1$. 
%
We wish to obtain a bound on $P^\dagger (l(v,\xi) \le 0)$ in terms of the corresponding probability under the empirical
probability mass function 
$P^0$. 
Note that, since the $v_h$ are real-valued, any uniform bound on the worst performance of \emph{any} $F$ cannot be stated as cleanly as in Proposition~\ref{prop:gen_uc_bounds}(a).
We therefore introduce an additional parameter $\Delta >0$
in
the statement of Proposition~\ref{prop:gen_uc_bounds}(b).
The parameter $\Delta$ 
plays an important role in the analysis. 
The upper bound described above becomes increasingly lax as $\Delta\searrow 0$. It is not merely a convenient device to ease the construction of the rigorous guarantee, but it also has an important practical implication in the sparse ensembling formulation~\eqref{opt:sparse_subprob}. Our experiments show that the $\Delta$ in this formulation, which plays the same role as in Proposition~\ref{prop:gen_uc_bounds}(b), must be a small positive value for the optimal sparse convex combinations identified by the formulation to be of higher generalization power than the individual DNF models.

\textbf{Proposition~\ref{prop:gen_uc_bounds}(b).}
\textit{
For any convex ruleset combination $F = \sum_k v_k h_k$, define $\bbC'\;\DefAs\; \max_k \{c(h_k)\}$ to be the maximum complexity of its constituent conjunctive terms, and let the predicted label be $\hat{y} = \ndctr{F(x) \ge 1/2}$. Measure the misclassification loss using $l(v,\xi)$ defined in~\eqref{def:binclass_loss}. Then, for any given $\delta\in(0,1)$, with probability at least $1-\delta$, we have
\begin{align} \tag{\ref{prop_eq:gen_F}}
		P^{\dagger} (l(v,\xi) \le 0) & \; \le \; P^{0} (l(v,\xi) \le \Delta) +
	O\bigg( \frac 1 {\sqrt{N}} \bigg(\frac {(\bbC' -1)}{\Delta^2} 
 \log N  + \log \frac 1 {\delta}\bigg)^{1/2}\bigg).
\end{align}
}
%


\noindent\textbf{Proof of Proposition~\ref{prop:gen_uc_bounds}(b).}
We start by noting that the set of convex weights $\{v_h,\,h\in\cH=\cH(\bbC')\}$ themselves represent a 
probability mass function. 
The analysis of the generalization error of $F$ utilizes this fact and approximates $F$ using an $M$-sized collection of DNFs sampled from $(v_h)$, as introduced by~\citet{sfbl98}. Define the approximation set 
\[
\cG_M := \bigg\{g\;:x\rightarrow\bigg\{0,\frac 1 M ,\frac 2 M, \ldots, 1\bigg\}\;\bigg|\; g(x) = \frac 1 M \sum_{m\in[M]} h_m(x),\;\mbox{where}\,\,\, h_m \in \cH(\bbC') , \,\,\, \forall \,m\in[M] \bigg\}.
\]
(We will specify the right size $M$ to use later in the proof.) Note that $|\cG_M| = |\cH|^M$. The analysis of $g\in\cG_M$ is easier than $F$ directly because $g$ only takes values in the lattice $\{0, 1/M, 2/M,\ldots, 1\}$. Let $\cV$ be the distribution induced on $\cG_M$ by sampling the components of a $g\in\cG_M$ from the 
probability mass function 
formed from the coefficients $(v_h)$ of $F$. 
Our main result is established in two steps. In the first step, we start by uniformly bounding the probability of the misclassification event $\{l(h,\xi) \le 0\}$ of $F$ under the true distribution $P^{\dagger}$ with the probability that $l(g,\xi) \le \Delta/2$ under the empirical 
probability mass function 
$P^0$ for an \emph{arbitrarily} chosen $g\in\cG_M$ that is sampled from the distribution $\cV$. We then bound the latter with the probability that $l(h,\xi) \le \Delta$ under the same empirical 
probability mass function 
$P^0$. 

Let $A=\{ l(h,\xi) \le 0\}$ and $B=\{ l(g,\xi)\le \Delta / 2\}$, where $g$ is any (fixed) element from $\cG_M$ and $\Delta>0$ is an arbitrary constant. Then, by the law of total probability and the definition of conditional probability, we have that $\pr_\dagger A \,\,=\,\, \pr_\dagger [A \cap B] \,\,+\,\, \pr_\dagger [ A \cap B^c] = \pr_\dagger [A | B]\,\pr_\dagger B \,\,+\,\, \pr_\dagger [B^c | A] \,\pr_\dagger A \,\,\le\,\, \pr_\dagger B \,\,+\,\, \pr_\dagger[ B^c | A]$. Thus, we conclude
\[
\pr_\dagger (l(h,\xi) \le 0) \;\le\; \pr_\dagger \bigg( l(g,\xi) \le \frac {\Delta}  2\bigg) \;+\; \pr_\dagger \bigg( l(g,\xi) > \frac {\Delta}  2 \,\,\bigg|\,\, l(h,\xi) \le 0\bigg).
\]
Since this applies to any $g\in\cG_M$, we can take expectations on the right hand side with respect to the distribution $\cV$ on $\cG_M$ to obtain that
\begin{align}
	\pr_\dagger ( l(h,\xi) \le 0) &\le \ex_{\cV} \bigg[ \pr_\dagger \bigg( l(g,\xi) \le \frac{\Delta} 2\bigg)\bigg] + \ex_\dagger \bigg[ \pr_{\cV} \bigg( l(g,\xi) > \frac {\Delta} 2  \,\,\bigg|\,\, l(h,\xi) \le 0\bigg)\bigg].\label{prf2:initUB}
\end{align}
We shall bound in turn the inner probability terms within the expectations of each summand on the right hand side of \eqref{prf2:initUB}. 
For the first summand, we treat the $g$ as a fixed sum of DNF terms, and thus apply the same union bound analysis from the proof of Proposition~\ref{prop:gen_uc_bounds}(a) to the Bernoulli random variable $\ndctr{l(g,\xi) \le \frac {\Delta} 2}$. This yields that the probability (over the sampling of $\cD$ from $P^{\dagger}$) that there exists at least one $g$ and $\Delta>0$ for which 
\begin{equation}
\pr_\dagger \bigg( l(g,\xi) \le \frac{\Delta} 2\bigg) > \pr_{0} \bigg( l(g, \xi) \le \frac {\Delta} 2 \bigg) \,\,+\,\, \lambda_M
    \label{prf2:bndxx}
\end{equation}
is at most $$(M+1) |\cH|^M \exp(-2N \lambda^2_M).$$ The term $\exp(-2N\lambda^2_M)$ comes from the Chernoff bound for \emph{any} fixed $g$ and $\Delta$. There are $|\cG_M| = |\cH|^M$ such $g$ over which to take the union. Moreover, each $g\in\cG_M$ takes one of only $M+1$ distinct values $\{0,1/M, \ldots, 1\}$, and there are at most $(M+1) |\cH|^M$ combinations of $g$ and $\Delta$ to consider. 
Hence, for any fixed $M$, we proceed as in the proof of Proposition 1 to set 
\begin{align}
\label{prf2:choosen_lambda}
\lambda_M = \bigg(\frac 1 {2N} \log \bigg(\frac{(M+1) |\cH|^M}{\delta_M}\bigg)\bigg)^{\frac{1}{2}},
\end{align}
and thus obtain that, with at least probability $1- \delta_M$, the bound in~\eqref{prf2:bndxx} is satisfied
for \emph{every} $g$ and $\Delta$. Taking the expectation needed in~\eqref{prf2:initUB} with respect to the distribution $\cV$, we have with probability $1-\delta_M$ that
\[
\pr_{\dagger,\cV} \bigg( l(g,\xi) \le \frac {\Delta} 2\bigg) \le \pr_{0,\cV} \bigg( l(g,\xi) \le \frac {\Delta} 2 \bigg) + \lambda_M.
\]
To complete the analysis of this first summand of~\eqref{prf2:initUB}, we need to provide an upper bound for the first term on the right hand side of the above expression. 
Utilizing the same conditioning argument that led to~\eqref{prf2:initUB}, we can write
\begin{equation}
\pr_{0,\cV} \bigg( l(g,\xi) \le \frac {\Delta} 2\bigg) \;\le\; \pr_{0} \bigg( l(h,\xi) \le {\Delta} \bigg)\,\, +\,\, \ex_{0} \bigg[ \pr_{\cV} \bigg( l(h,\xi) > {\Delta} \,\,\bigg|\,\, l(g,\xi) \le \frac {\Delta} 2 \bigg)\bigg].\label{prf2:nexted_bnd}
\end{equation}

Here, the $F$ in the first term on the right hand side is unaffected by the distribution $\cV$ of $g\in\cG_M$.  The probability inside the expectation in the second term is similar in spirit to that of the second summand in~\eqref{prf2:initUB} and deals with the sampling of $g\in \cG_M$ from the induced distribution $\cV$. Note that $F(x) = \ex_{\cV} g(x)$ and the 
probability mass function 
$(v_h)$ from which the $g(x)\sim\cV$ is sampled has finite support of size $|\cH|$. This allows us to exploit the Chernoff bound for finite support distributions. 
In particular, for the second term in~\eqref{prf2:nexted_bnd}, we obtain
\begin{align}
    \pr_{\cV} \bigg( l(h,\xi) > {\Delta} \,\,\bigg|\,\, l(g,\xi) \le \frac {\Delta} 2 \bigg) \;\le\; \exp\bigg(-\frac{M\Delta^2} 8\bigg).\label{prf2:2nd_2}
\end{align}
For the second summand in~\eqref{prf2:initUB},  we analogously have
\begin{align}
\pr_{\cV} \bigg ( l(g,\xi) > \frac {\Delta} 2\,\,\bigg|\,\, l(v,\xi) \le 0\bigg) \;\le\; \exp\bigg(-\frac {M \Delta^2} 8\bigg). \label{prf2:2nd_1}
\end{align}
Combining~\eqref{prf2:initUB},~\eqref{prf2:bndxx},~\eqref{prf2:nexted_bnd},~\eqref{prf2:2nd_2} and~\eqref{prf2:2nd_1}, we have established, for an arbitrary $M$ and with probability at least $1-\delta_M$, that
\begin{align}\label{prf2:final_ish}
\pr_\dagger ( l(h,\xi) \le 0) \;\;\le\;\;\pr_{0} \bigg( l(h,\xi) \le {\Delta} \bigg)\,\, +\,\, 2 \exp\bigg(-\frac {M \Delta^2} 8\bigg) + \lambda_M,
\end{align}
where $\lambda_M$ is as chosen in~\eqref{prf2:choosen_lambda}. To obtain a bound that applies uniformly for \emph{any} $M$, we once again appeal to the union bound analysis by setting $\delta_M = \delta \Big( \frac{1}{M} - \frac{1}{M+1}\Big)$. Thus, with probability at least $1 - \sum_{M\ge 1} \delta_M = 1 - \delta$, we have that the bound in~\eqref{prf2:final_ish} is uniformly satisfied for any $M\ge 1$. 

There are two terms that depend on $M$ in the right side of~\eqref{prf2:final_ish}. The middle term exponentially decreases with $M$, while $\lambda_M$ (with the form given in~\eqref{prf2:choosen_lambda} after substituting $\delta_M$) has a leading term that increases as a small polynomial of $M$. In order to strike a balance between these terms, 
we choose a value for $M$ by setting $\exp(-M\Delta^2/8) = \sqrt{\log |\cH| /N}$. Thus, we obtain
$$M = \frac 4  \Delta^2 \log\bigg(\frac N {\log|\cH (\bbC')|}\bigg)  .$$
Finally, using the lower bound on $\log |\cH(\bbC')|$ from~\eqref{bounds:size_H} yields the desired result. 
\hfill\ProofEnd

\section{Column Generation for Step 1 in Sparse Ensembling Algorithm}\label{apdx:cg-details} 

Our column generation implementation is an extension of the approach of~\cite{dgw18}.
We refer to that paper for additional details beyond those presented here for our extension of the approach.

The size of the set $\cK$ of conjunctions (rules) in the formulation to search over the space of rulesets $\cH(\bbC')$ clearly grows as an exponential function of the number of binary features.
Solving 
IPs
with exponentially many variables is impractical; even the associated linear program relaxation is hard to solve.
Notice that the number of constraints (which are not variable constraints) in this formulation is bounded by the number of data points with label 1. 
One approach to solving large linear programs that have few constraints but very many variables is to use {\em column generation}.
This approach is motivated by the following: (1) there is a solution to the linear program in which the number of nonzero variable values is bounded above by the number of constraints; and (2) when solving the linear program relaxation with the simplex method, it is fairly common that the total number of variables that are ever set to nonzero values in intermediate steps is a small multiple of the number of constraints.
In other words, only a small subset of all possible $w_k$ variables (rules) is generated explicitly and the optimality of the linear program is guaranteed by iteratively solving a {\em pricing problem} that implicitly checks the dual constraints.

To apply this framework to the 
IP
(which we refer to as the ``master integer program'' or MIP, with associated linear program relaxation denoted by  ``master linear program'' or MLP), we first replace $\cK$ with a very small subset of its elements and explicitly solve the linear program relaxation of the resulting smaller problem, which we call the {\em Restricted master linear program} or RMLP. 
Any optimal solution of the RMLP can be extended to a solution of the MLP with the same objective value by setting all missing $w_k$ variables to zero; this solution provides an upper bound on the optimal solution value of the MLP.
    One can potentially improve such a solution by augmenting the RMLP with additional variables corresponding to some of the missing rules. 
The second step is to identify such rules without explicitly considering all of them. 
Repeating these steps until there are no improving rules (i.e.,~variables missing from the RMLP that can reduce the cost) solves the MLP to optimality.

To find the missing improving rules, one needs to check for variables in the MLP missing from the RMLP that  have negative reduced cost (with respect to the dual solution from the RMLP applied to variables in the MLP not present in the RMLP).
     The reduced cost of a missing variable (or variable set to zero in the MLP) yields the maximum possible change in objective value per unit increase in the value of that variable. Therefore, if all missing variables from the RMLP have nonnegative reduced cost, then the optimal solution for the RMLP yields an optimal solution of the MLP.
     Furthermore, missing variables that have large negative reduced costs are more likely to improve the objective value of the RMLP.   We next formulate an optimization problem that uses the optimal dual solution of the RMLP and checks for negative reduced cost columns in the MLP. Let  $\mu_n \geq 0$ denote the dual variable associated with constraint \eqref{master-const1} corresponding to the $n$th data point with label 1. Let $\lambda \geq 0$ be the dual variable associated with the complexity bound constraint. Let $\delta_n \in \{0,1\}$ denote whether the $n$th sample satisfies rule $k$.  Recall that $c_k$ is the complexity of rule $k$, and then its  reduced cost  is equal to 
\begin{equation}\label{eqn:reducedCost}
\sum_{n: y_n=0} P_n \delta_n - \sum_{n: y_n = 1} P_n \mu_n \delta_n + \lambda c_k.
\end{equation}
The first term in \eqref{eqn:reducedCost} is the cost of rule $k$ in the objective function \eqref{master-obj}, expressed in terms of $\delta_n$. The second term is the sum of the dual variables associated with constraints \eqref{master-const1}. The last term is the dual variable associated with the complexity constraint  multiplied by the complexity of rule $k$.

We now formulate an 
IP
to express rules as conjunctions of the original features.
Let the decision variable $z_j\in\{0,1\}$ denote if feature $j$ is selected in the rule. 
Let $S_n$ correspond to the zero-valued features 
in sample $n$.
Recall that there are $\ell$ features and $[\ell]$ forms an index set over them.
Then the {\em Pricing Problem} below identifies the rule $k$ missing from the RMLP that has the lowest reduced cost:
\begin{subequations}
\begin{align}
z_{CG}~=~&\textbf{min} \quad
\lambda\Biggl(1 + \sum_{j\in J} z_j\Biggr) - \sum_{n: y_n=1}P_n\mu_n \delta_n +\sum_{n: y_n = 0} P_n \delta_n &\qquad\qquad\qquad\label{eqn:pip_obj}\\
&~~~\textbf{s.t.} \qquad
  \delta_n +z_j \leq 1, \qquad\qquad\;\; j \in S_n, \; \forall n: y_n = 1 , \label{eqn:pip_Perr}\\ 
&~~~\qquad\quad \delta_n \geq 1-\sum_{j\in S_n} z_j, \quad\qquad \delta_n \geq 0, \; \forall n: y_n = 0 , \label{eqn:pip_Zerr}\\
&~~~\qquad\quad\sum_{j\in [\ell]} z_j \leq c\;;
\qquad\qquad\;\;
z_j  \in\{0,1\},  \;\;\;
j\in [\ell]. 
\qquad\qquad 
\label{eqn:pip_card}
\end{align}
\end{subequations}
The first term in \eqref{eqn:pip_obj} expresses the complexity $c_k$ in terms of the number of selected features.  Constraints \eqref{eqn:pip_Perr}, \eqref{eqn:pip_Zerr} ensure that the rule acts as a conjunction, i.e.,~it is satisfied ($\delta_n = 1$) only if no zero-valued features are selected ($z_j = 0$ for $j \in S_n$).  Similar to $\xi_n$ in the MIP, the variables $\delta_n$ do not have to be explicitly defined as binary due to the objective function.
Constraint \eqref{eqn:pip_card} bounds the number of features allowed in any rule. 

The optimal solution to  the Pricing Problem above provides the rule with the minimum reduced cost that is missing from the RMLP.
The reduced cost of this rule equals $z_{CG} $ and if $z_{CG} <0$, then the corresponding variable is added to the RMLP. 
More generally, any feasible solution to the Pricing Problem that has a negative objective function value provides a rule with a negative reduced cost and therefore can be added to the RMLP to improve its value.

\section{Details on Numerical Experiments and Additional Results} \label{apdx:expts}
The datasets were preprocessed following the same approach as described by~\citet{dgw18}. We used the standard “dummy”/“one-hot” coding to transform categorical features into multiple binary features. Each category is represented by two binary features, one indicating the presence of that category and one indicating the absence of that category.
The non-binary numerical features are transformed into a sequence of binary features where each feature represents the satisfaction or not of a certain threshold and its negation. As an example, for a certain numerical feature $X$, we could create the features $X \leq 1$, $X \leq 2$ and $X > 1$, $X > 2$. The thresholds used were deciles computed from the samples.

Two of the UCI repository datasets were slightly modified as follows:
\begin{itemize}
    \item For the liver dataset, we used the number of drinks as the output variable instead of the selector variable. The number of drinks $Y$ was represented by the two features $Y \leq 2$ and $Y > 2$.
    \item For the heart dataset, we used only the Cleveland data and removed 4 samples with ‘ca’ = ?, yielding 299 samples. The label was binarized as either $X = 0$ or $X > 0$.
\end{itemize}

The FICO dataset contains $23$ numerical features of the credit history of $10,459$ individuals ($9871$ after removing records with all entries missing) for predicting repayment risk (good/bad). 
We note that the FICO dataset was also slightly modified due to missing values and the presence of special values.
In particular, the 588 records that only contain missing values (represented by -9) were removed. Entries meaning `no inquiries or delinquencies observed' (represented by -7) were replaced by the maximum number of elapsed months in the data plus 1. The remaining missing values (represented by -9) and entries meaning `not applicable' (represented by -8) were combined into a single null category.
{During binarization, a special indicator was created for these null values and all other comparisons
with the null value return False.}
Values greater than 7 (other) in ‘MaxDelq2PublicRecLast12M’ were
imputed as 7 (current and never delinquent) based on the corresponding values in ‘MaxDelqEver’.

The rest of this section expands on the experimental results of our DR (Distributionally robust Rulesets) algorithm. 
Subsection~\ref{ssec:apdx-expt-vscg} compares the performance characteristics of our sparse convex ensembling algorithm to the IP-based sparsifier proposed by~\citet{dgw18}. We do this both along the criterion of computational budget required by each method in Table~\ref{tab:compute} and the test performance in Table~\ref{tab:diff-CG} attained when each method is forced to obey the same complexity constraint.
In Subsection~\ref{ssec:apdx-expt-dr}, we present a number of tables documenting the results of numerical experiments where specific parameters of the DR algorithm are varied. Subsection~\ref{ssec:apdx-expt-drsp} provides plots of the performance of the trajectories of the sparse ensembling algorithm for a selection of dataset instances. 
%
All Results are presented with a Mean and a $95\%$ Confidence Interval, as in the main body of the paper. The result with the best mean performance value among the two columns are highlighted in \textbf{bold}.

\subsection{Comparison to IP Method of~\citet{dgw18}.} \label{ssec:apdx-expt-vscg}

\begin{table*}[!tbhp]
\caption{Comparison of Computation Times Between our DR Approach and the CG Approach to Solve One Training Run for seven UCI Repository Datasets and the FICO Dataset, Averaged over ten Partition Permutations.
All Results presented with a Mean and a $95\%$ Confidence Interval. 
}
\begin{center}
\begin{tabular}{|l||r||r||c|}
\hline
Dataset &\multicolumn{2}{c||}{Computation Time} & Relative  DR \\
Name 
& {DR} 
& {CG} 
& Improvement \\
\hline
heart	&1431.73(17.91)	&6015.66(115.65)	
&4.2\\
ILPD	&2136.25(25.86)	&11868.06(333.90)
&5.6\\
FICO	&9548.59(98.22)	&42250.38(691.78)
&4.4\\
ionosphere	&6008.45(16.12)	&23168.82(093.65)
&3.9\\
liver	&1661.16(12.84)	&6472.05(093.28)
&3.9\\
pima	&5250.25(18.00)	&22827.18(116.23)
&4.3\\
transfusion	&775.87(08.03)	&3853.68(060.24)
&5.0\\
WDBC	&5516.39(35.64)	&22485.30(212.41)
&4.1\\
\hline
\end{tabular}
\end{center}
\label{tab:compute}
\end{table*}

Table~\ref{tab:compute} presents the computation time results for our DR approach and the CG approach to solve one training run for each of the eight datasets.
The mean computation time is provided for both approaches averaged over $10$ partition permutations, together with the corresponding confidence intervals and the relative reduction in computation time provided by our DR approach in comparison with the CG approach.
We observe that DR provides significant computational improvements over CG, with relative improvements ranging from $3.9$ to $5.6$.

\begin{table*}[htb]
\caption{Results for CG when Complexity Observed in DR Results are Imposed as Constraints on CG. 
}
\begin{center}
\begin{tabular}{|l||c|c||c|c||}
\hline
Dataset 
&\multicolumn{2}{c||}{Test Performance (\%)}
&\multicolumn{2}{c||}{Model Complexity} \\
Name 
&DR&CG
&DR&CG \\
\hline
heart 
& \textbf{81.0}(4.4) & 78.9(2.7)
& 22.0(4.2) & \textbf{21.7}(4.1) \\
ILPD 
& \textbf{71.7}(2.4) & 70.6(2.0)
& \textbf{10.5}(0.9) &  \textbf{10.5}(0.9) \\
FICO 
& \textbf{71.8}(0.6) & 70.9(0.6)
& 21.5(3.9) & \textbf{21.2}(4.2) \\
ionosphere 
& \textbf{93.2}(3.5) & 88.5(3.8)
& 26.5(3.4) & \textbf{25.2}(3.5) \\
liver 
& \textbf{62.0}(2.8) & 57.9(5.3)
& 29.0(1.9) & \textbf{28.8}(1.8) \\
pima 
& \textbf{76.7}(2.8) & 73.3(3.3)
& 22.0(4.2) & \textbf{21.6}(4.5) \\
transfusion 
& \textbf{79.0}(2.1) & 77.9(1.7)
& 13.0(1.5) & \textbf{12.0}(1.7) \\
WDBC 
& \textbf{94.9}(1.6) & 94.0(1.4)
& 29.0(1.9) & \textbf{28.1}(1.8) \\
\hline
\end{tabular}
\end{center}
\label{tab:diff-CG}
\end{table*}

Table~\ref{tab:diff-CG} presents another illustration of the better generalization of the ensembles constructed by the DR method than the DNF models produced by the CG method. Test performance of the CG and DR methods are provided where the total complexity budget $\bbC$ is kept fixed to the value determined by the DR method when run with the settings that produced the results in Table~\ref{tab:perf-comp}. In every case, we observe that the test performance produced by the DR method exceeds (by a considerable margin in some cases) that produced by the CG method for nearly the same model complexity.

\subsection{Varying Parameters of DR Method} \label{ssec:apdx-expt-dr}

In each table provided below, we present two sets of columns each for the classification performance over unseen test data, for the complexity of models constructed, and for the run length (number of iterations $n$) of the DR method. The first column provides the results of the DR method with hyperparameters as presented in Table~\ref{tab:perf-comp}, whereas the results of the second column are obtained by changing one of the hyperparameters at a time.

\begin{table*}[htb]
\caption{Sparse Ensembles Constructed from Low ($\bbC'=5$) Complexity DNF Models are Compared to Ensembles Constructed from Moderate ($\bbC'=10$) Complexity DNF Models. 
}
\begin{center}
\resizebox{\textwidth}{!}{%
\begin{tabular}{|l||c|c||c|c||c|c|}
\hline
Dataset 
&\multicolumn{2}{c||}{Test Performance (\%) of DR}
&\multicolumn{2}{c||}{Model Complexity of DR}
&\multicolumn{2}{c|}{$n$, number of Iterates of DR} \\
Name 
&$\bbC'=5$&$\bbC'=10$
&$\bbC'=5$&$\bbC'=10$
&$\bbC'=5$&$\bbC'=10$ \\
\hline
heart 
& \textbf{81.0}(4.4) & 79.6(5.4)
& \textbf{22.0}(4.2) & 30.0(0.0)
& \textbf{31.7}(4.2) & 35.2(4.1) \\
ILPD 
& \textbf{71.7}(2.4) & 69.8(3.2)
& \textbf{10.5}(0.9) &  26.0(3.0)
& \textbf{26.0}(0.0) & 26.4(0.4) \\
FICO 
& \textbf{71.8}(0.6) & 71.3(0.9)
& \textbf{21.5}(3.9) & 29.0(1.9)
& 36.2(5.3) & \textbf{33.2}(4.4) \\
ionosphere 
& \textbf{93.2}(3.5) & 93.0(2.4)
& \textbf{26.5}(3.4) & 30.0(0.0)
& 45.1(3.7) & \textbf{37.1}(5.4) \\
liver 
& \textbf{62.0}(2.8) & 60.4(6.8)
& \textbf{29.0}(1.9) & 30.0(0.0)
& 38.1(5.2) & \textbf{32.9}(5.5)\\
pima 
& \textbf{76.7}(2.8) & 75.4(3.5)
& \textbf{22.0}(4.2) & 30.0(0.0)
& \textbf{29.4}(2.5) & 38.4(4.9) \\
transfusion 
& \textbf{79.0}(2.1) & 78.9(2.0)
& \textbf{13.0}(1.5) & 30.0(0.0)
& \textbf{28.9}(2.5) & 30.2(3.1)\\
WDBC 
& 94.9(1.6) & \textbf{96.1}(1.5)
& \textbf{29.0}(1.9) & 30.0(0.0)
& \textbf{40.0}(4.5) & 44.9(4.2) \\
\hline
\end{tabular}
}
\end{center}
\label{tab:diff-bbcdash}
\end{table*}

The results presented in Table~\ref{tab:diff-bbcdash} contrast the ensembling of DNF models with a low value for $\bbC'=5$ to the ensembling of DNF models with double the complexity $\bbC'=10$. All other parameters are fixed to their value used in Table~\ref{tab:perf-comp}: $P=P^0$ in the ensembling formulation~\eqref{opt:sparse_subprob}, $\bbC$ is cutoff at $30$, the DRO formulation~\eqref{opt:ensemble_rob_fmln} sets $\rho=0.05$, and the outer iterations stop when no progress is made in $20$ consecutive iterations. From the results, it is clear that the low complexity DNF models grant more generalization flexibility to the sparse ensembling formulation~\eqref{opt:sparse_subprob}, beating the best ensembles assembled from $\bbC'=10$ in six of the eight cases (and significantly so, such as for the `liver' dataset instance) and are within the $95\%$ confidence interval of the best in all cases. Moreover, the total model complexity $\bbC$ produced for $\bbC'=5$ ensembles are significantly smaller in all instances. The generalization guarantee provided by Proposition~\ref{prop:gen_uc_bounds}(b) anticipates this, since it predicts that  performance may degrade as $\bbC'$ increases.

\begin{table*}[htb]
\caption{Aggregate Complexity of Sparse Ensemble Determined by~\eqref{opt:sparse_subprob} with a Cutoff at $\bbC\le 30$ Compared to a Cutoff at $\bbC\le 60$}
\begin{center}
\resizebox{\textwidth}{!}{%
\begin{tabular}{|l||c|c||c|c||c|c|}
\hline
Dataset 
&\multicolumn{2}{c||}{Test Performance (\%) of DR}
&\multicolumn{2}{c||}{Model Complexity of DR}
&\multicolumn{2}{c|}{$n$, number of Iterates of DR} \\
Name 
&cutoff $=30$& cutoff $=60$
&cutoff $=30$& cutoff $=60$
&cutoff $=30$& cutoff $=60$ \\
\hline
heart 
& \textbf{81.0}(4.4) & \textbf{81.0}(4.1)
& \textbf{22.0}(4.2) & 38.5(8.5)
& \textbf{31.7}(4.2) & 36.7(5.1) \\
ILPD 
& \textbf{71.7}(2.4) & \textbf{71.7}(2.4)
& \textbf{10.5}(0.9) & \textbf{10.5}(0.9)
& \textbf{26.0}(0.0) & \textbf{26.0}(0.0) \\
FICO 
& \textbf{71.8}(0.6) & \textbf{71.8}(0.7)
& \textbf{21.5}(3.9) & 24.5(7.1)
& \textbf{36.2}(5.3) & 38.8(5.8) \\
ionosphere 
& \textbf{93.2}(3.5) & 92.9(3.7)
& \textbf{26.5}(3.4) & 29.5(6.0)
& \textbf{45.1}(3.7) & 47.2(2.3) \\
liver 
& \textbf{62.0}(2.8) & 61.7(2.9)
& \textbf{29.0}(1.9) & 39.0(7.6)
& \textbf{38.1}(5.2) & 39.0(4.5)\\
pima 
& \textbf{76.7}(2.8) & \textbf{76.7}(2.8)
& \textbf{22.0}(4.2) & \textbf{22.0}(4.2)
& \textbf{29.4}(2.5) & \textbf{29.4}(2.5) \\
transfusion 
& \textbf{79.0}(2.1) & \textbf{79.0}(2.1)
& \textbf{13.0}(1.5) & \textbf{13.0}(1.5)
& \textbf{28.9}(2.5) & \textbf{28.9}(2.5)\\
WDBC 
& 94.9(1.6) & \textbf{95.2}(1.7)
& \textbf{29.0}(1.9) & 40.0(5.5)
& \textbf{40.0}(4.5) & 43.6(4.1) \\
\hline
\end{tabular}
}
\end{center}
\label{tab:diff-cutoff}
\end{table*}

Table~\ref{tab:diff-cutoff} varies the cutoff allowed in the inner sparse ensembling iterations where $\bbC$ is increased as $\bbC= m \cdot \bbC',\,\,m=2,3,\ldots$. The results presented in Table~\ref{tab:perf-comp} of the main body cuts off at $\bbC\le 30$ or earlier, which is compared here to a larger cutoff of $60$. Other parameters are fixed to be the same as in Table~\ref{tab:perf-comp}: $\bbC'=5$, the formulation~\eqref{opt:sparse_subprob} uses $P=P^0$, the DRO formulation~\eqref{opt:ensemble_rob_fmln} sets $\rho=0.05$, and outer iterations stop when $20$ consecutive iterations do not improve the training performance. The generalization performance is similar for both values of the cutoff but, as can be expected, the higher cutoff has significantly more complex (and harder to interpret) models.
Overall the cutoff of $30$ appears to be preferable from the viewpoint of all three factors of good generalization performance at a low computational cost while producing models with good interpretability.

\begin{table*}[htb]
\caption{DRO Radius Parameter $\rho$ is varied from $0.05$ to $0.5$}
\begin{center}
\resizebox{\textwidth}{!}{%
\begin{tabular}{|l||c|c||c|c||c|c|}
\hline
Dataset
&\multicolumn{2}{c||}{Test Performance (\%) of DR}
&\multicolumn{2}{c||}{Model Complexity of DR}
&\multicolumn{2}{c|}{$n$, number of Iterates of DR} \\
Name
&$\rho=0.05$&$\rho=0.5$
&$\rho=0.05$&$\rho=0.5$
&$\rho=0.05$&$\rho=0.5$\\
\hline
heart 
& 81.0(4.4) & \textbf{81.9}(5.3)
& \textbf{22.0}(4.2) & 30.0(0.0)
& \textbf{31.7}(4.2) & 38.2(3.7) \\
ILPD 
& 71.7(2.4) & \textbf{71.8}(2.3)
& \textbf{10.5}(0.9) & \textbf{10.5}(0.9)
& \textbf{26.0}(0.0) & \textbf{26.0}(0.0) \\
FICO 
& \textbf{71.8}(0.6) & 71.6(0.8)
& 21.5(3.9) & \textbf{17.5}(4.4)
& 36.2(5.3) & \textbf{31.8}(2.6) \\
ionosphere 
& 93.2(3.5) & \textbf{93.5}(3.9)
& \textbf{26.5}(3.4) & 30.0(0.0)
& \textbf{45.1}(3.7) & 45.5(5.6) \\
pima 
& \textbf{76.7}(2.8) & \textbf{76.7}(3.6)
& \textbf{22.0}(4.2) & 25.5(3.5)
& \textbf{29.4}(2.5) & 44.1(3.6) \\
transfusion 
& 79.0(2.1) & \textbf{79.3}(2.4)
& 13.0(1.5) & \textbf{12.0}(1.5)
& 28.9(2.5) & \textbf{27.8}(2.5)\\
WDBC 
& 94.9(1.6) & \textbf{95.3}(1.0)
& \textbf{29.0}(1.9) & \textbf{29.0}(1.9)
& \textbf{40.0}(4.5) & 43.2(4.9) \\
\hline
\end{tabular}
}
\end{center}
\label{tab:diff-rho}
\end{table*}

In Table~\ref{tab:diff-rho}, results are presented for a larger DRO parameter $\rho=0.5$, keeping all other parameters the same as in Table~\ref{tab:perf-comp}: $\bbC'=5$, $\bbC$ is cutoff at $30$, the ensembling formulation~\eqref{opt:sparse_subprob} uses $P=P^0$, and iterations $n$ stop after $20$ consecutive iterations of little improvement. 
Recall that the parameter $\rho$ was set
according to the recommendation of Theorem~\ref{thm:dromain} to have $\rho = \chi^2_{{D_\eff +1 }, 1-\tilde{\alpha}} / N$, which under the choice of $\alpha=0.05$ and for the test datasets considered renders $\rho$ to be in $O(10^{-2})$ to $O(10^{-1})$, and thus we set $\rho=0.05$.
The results presented in Table~\ref{tab:diff-rho} illustrate that $\rho=0.5$ might provide improvements on test performance on some datasets, but this comes at the cost of slightly higher model complexity and computational cost (as represented by the number $n$ of DR iterates needed). Additionally, we found that the larger $\rho=0.5$ makes the DRO formulation too conservative for the `liver' dataset instance, and thus such results are omitted in Table~\ref{tab:diff-rho}.

\begin{table*}[htb]
\caption{Minimum Number of Consecutive Iterations of No-Improvement to Observe in Step 4 of Sparse Ensembling Algorithm before Stopping Outer Iterations: Test Performance, Model Complexity and number $t$ of Outer Iterations}
\begin{center}
\resizebox{\textwidth}{!}{%
\begin{tabular}{|l||c|c|c||c|c|c||c|c|c||}
\hline
Dataset
&\multicolumn{3}{c||}{Test Performance (\%) of DR}
&\multicolumn{3}{c||}{Model Complexity of DR} 
&\multicolumn{3}{c||}{$t$, number of Iterates of DR} \\
Name
&$10$&$20$&$30$
&$10$&$20$&$30$
&$10$&$20$&$30$\\
\hline
heart 
& 80.6(3.4) & \textbf{81.0}(4.4) & 80.0(4.4)
& 23.0(4.4) & \textbf{22.0}(4.2) & 25.0(3.9) 
& \textbf{20.5}(2.3) & 31.7(4.2) & 43.6(3.5) 
\\
ILPD 
& \textbf{71.7}(2.4) & \textbf{71.7}(2.4) & \textbf{71.7}(2.4)
& \textbf{10.5}(0.9) & \textbf{10.5}(0.9) & \textbf{10.5}(0.9)
& \textbf{16.0}(0.0) & 26.0(0.0) & 36.0(0.0) 
\\
FICO 
& \textbf{72.1}(0.7) & 71.8(0.6) & \textbf{72.1}(0.7)
& 23.0(3.7) & \textbf{21.5}(3.9) & 24.5(3.5)
& \textbf{21.8}(3.2) & 36.2(5.3) & 44.0(3.3) 
\\
ionosphere 
& 92.6(3.6) & 93.2(3.5) & \textbf{93.5}(3.5)
& \textbf{22.5}(4.6) & 26.5(3.4) & 28.0(2.5)
& \textbf{22.9}(3.7) & 45.1(3.7) & 49.1(1.7) 
\\
liver 
& \textbf{62.0}(3.3) & \textbf{62.0}(2.8) & 61.4(2.8)
& \textbf{28.0}(2.5) & 29.0(1.9) & 30.0(0.0)
& \textbf{24.6}(4.2) & 38.1(5.2) & 46.1(3.0) 
\\
pima 
& \textbf{77.1}(3.0) & 76.7(2.8) & 76.2(3.1)
& 26.0(3.9) & \textbf{22.0}(4.2) & \textbf{22.0}(4.2)
& \textbf{19.5}(2.4) & 29.4(2.5) & 39.2(2.7) 
\\
transfusion 
& \textbf{79.0}(2.1) & \textbf{79.0}(2.1) & \textbf{79.0}(2.1)
& \textbf{13.0}(1.5) & \textbf{13.0}(1.5) & \textbf{13.0}(1.5)
& \textbf{18.9}(2.5) & 28.9(2.5) & 38.9(2.5) 
\\
WDBC 
& 94.7(1.9) & 94.9(1.6) & \textbf{96.0}(1.2)
& 30.0(0.0) & \textbf{29.0}(1.9) & 30.0(0.0)
& \textbf{27.3}(3.7) & 40.0(4.5) & 47.4(2.1) 
\\
\hline
\end{tabular}
}
\end{center}
\label{tab:diff-stopcrit-A}
\end{table*}

Table~\ref{tab:diff-stopcrit-A} varies the stopping criterion parameter of how many consecutive iterations of  less than $0.5\%$ aggregate improvement (in training performance) to observe before terminating the outer iterations. We present results for this parameter being set at $10$ and $30$ and reproduce the results for $20$ as presented in the main body of the paper. The other parameters are fixed at their previous values: $\bbC'=5$, $\rho=0.05$, and $\bbC$ is cutoff at $30$. The more aggressive parameter value of $10$ can produce lower complexity models more quickly, while the more conservative setting of $30$ can provide improvements in test performance at the expense of longer computation time and less interpretable models. The value of $20$, chosen for the results presented in the main body, seems to attain a good balance over these three competing factors.

\subsection{DR Sample Paths}\label{ssec:apdx-expt-drsp}

\begin{figure}
    \centering
    \includegraphics[width=0.45\linewidth]{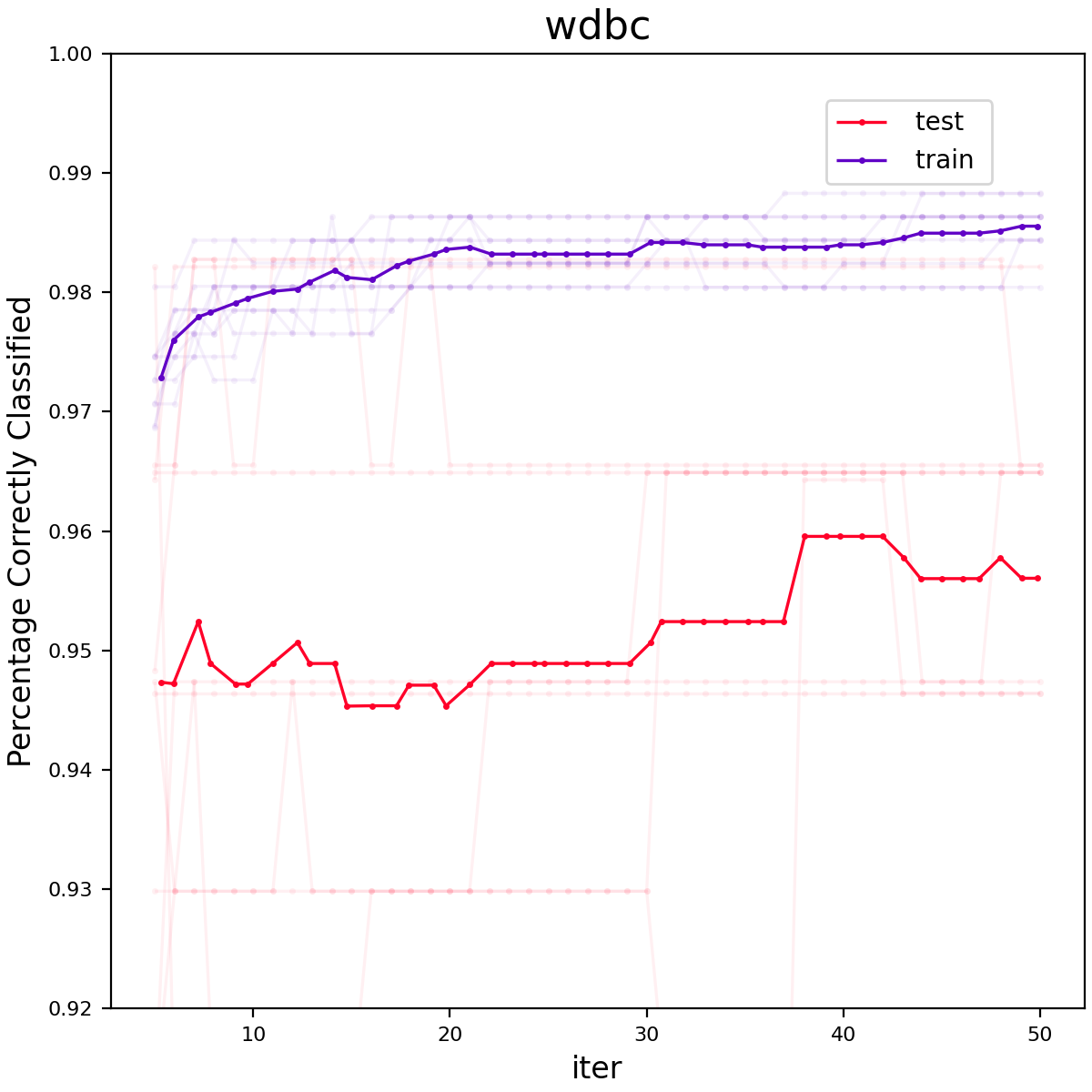}
    \includegraphics[width=0.45\linewidth]{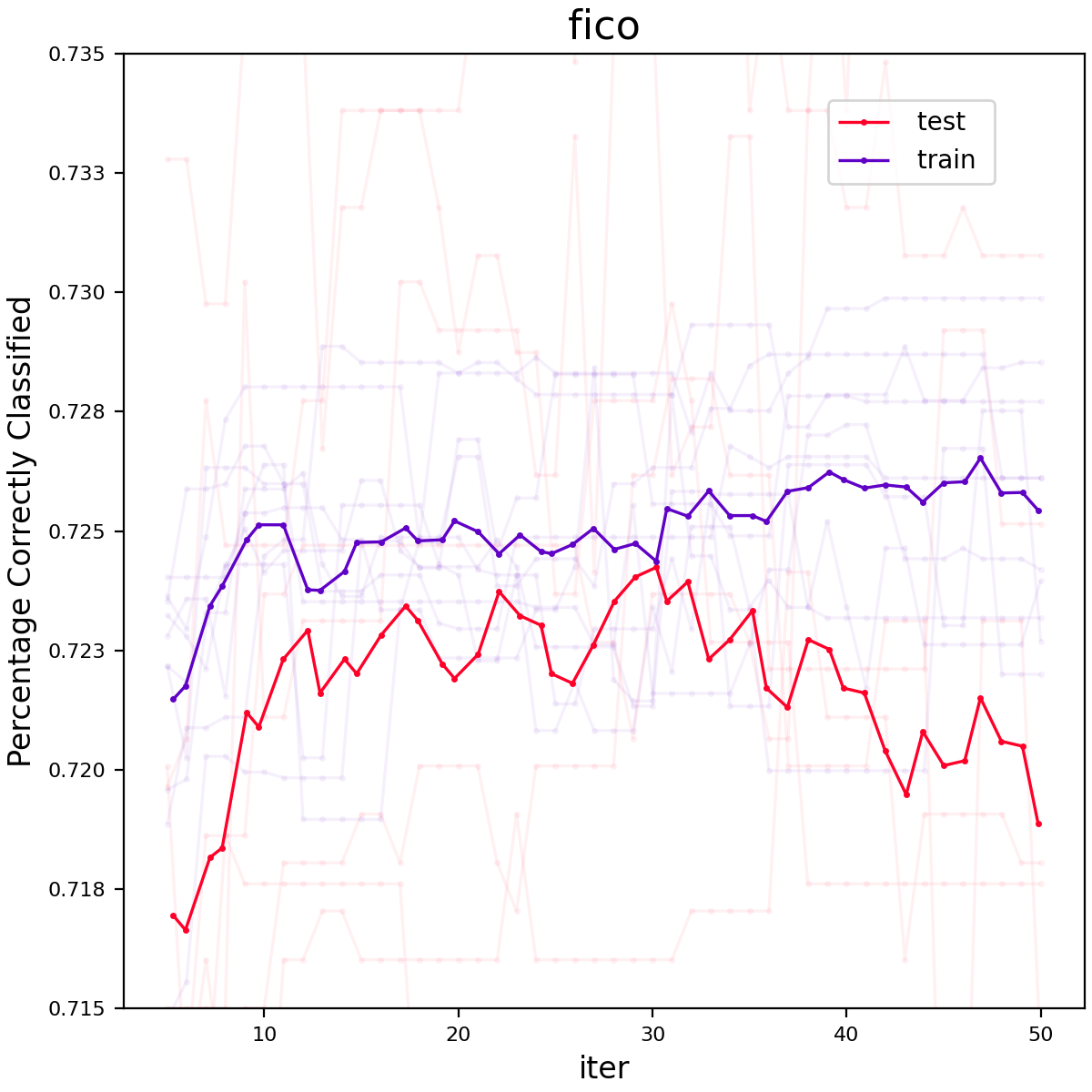}
    \includegraphics[width=0.45\linewidth]{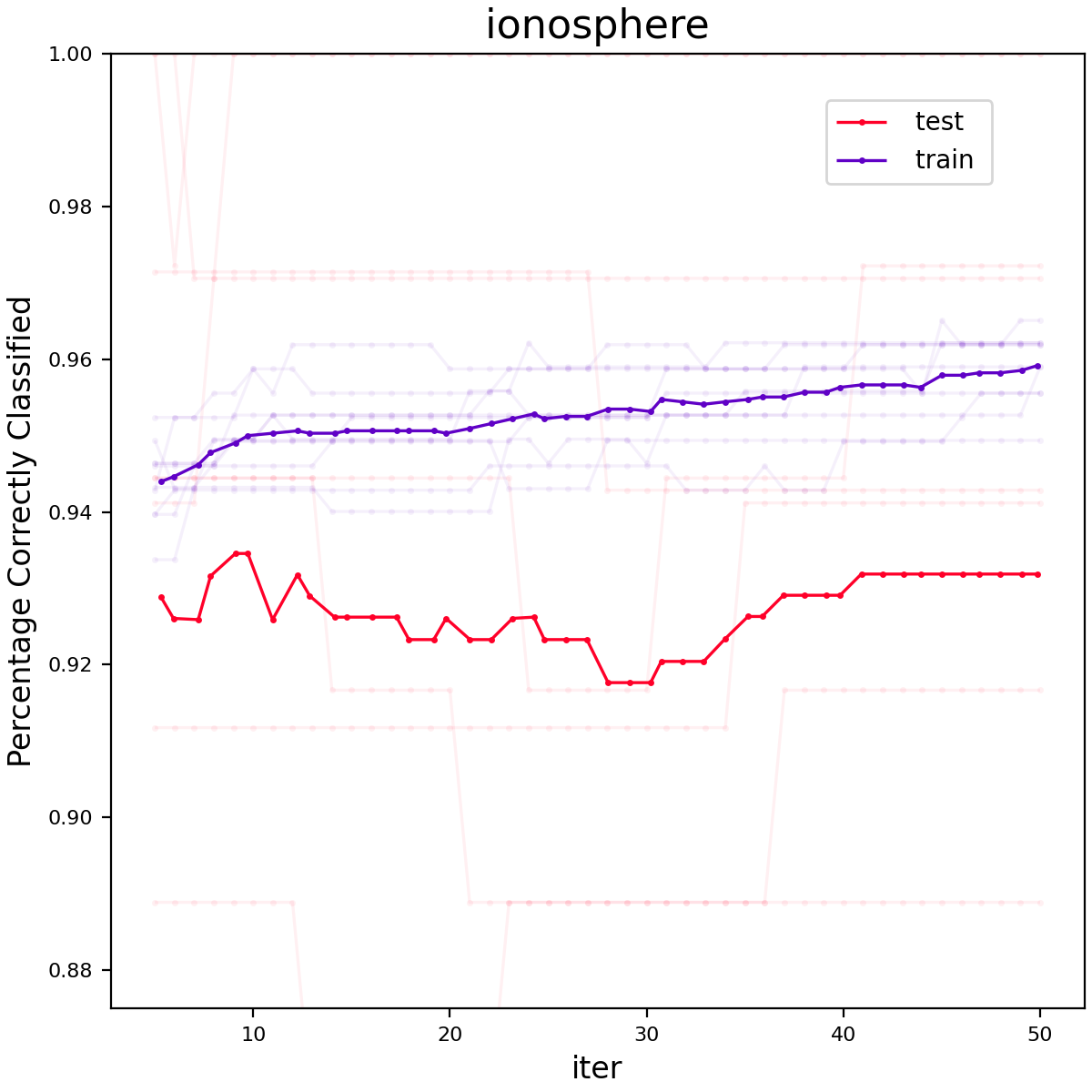}
    \includegraphics[width=0.45\linewidth]{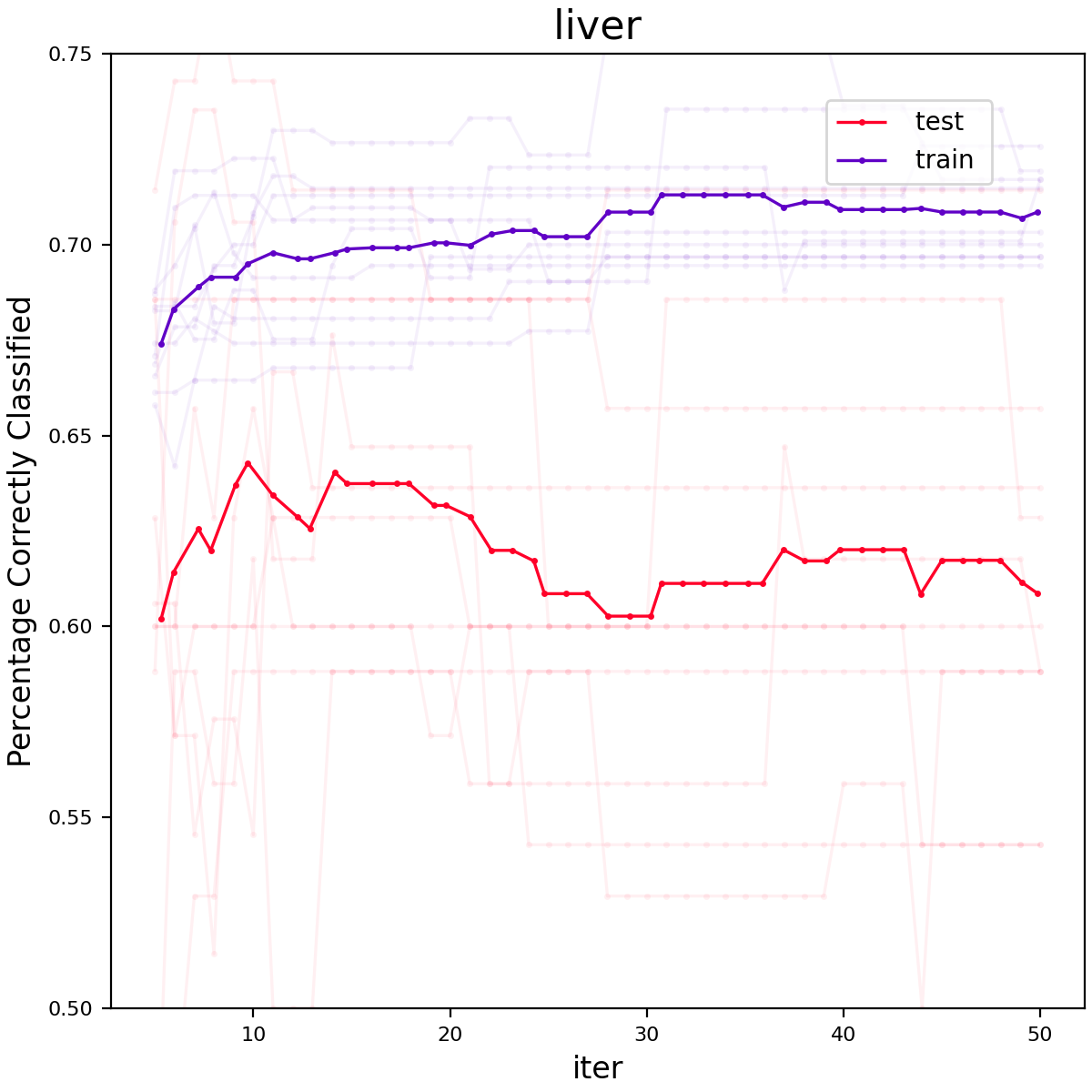}
    \caption{Iterate Trajectories of Percentage Correctly Classified in Test and Training Datasets by Sparse Ensembling Algorithm over four dataset instances.}
    \label{fig:enter-label}
\end{figure}

We present in Figure~\ref{fig:enter-label} the iterate trajectories for four datasets: WDBC, FICO, ionosphere and liver. For each dataset, the training and testing performance, in terms of the percentage that is correctly classified, is presented up to iteration $\ell=50$. The mean performance over the $10$ replications are also provided. 
Recall that, in each replication, our procedure terminates when 20 consecutive iterations do not improve the training performance by more than 0.5\%. 
In the WDBC and ionosphere cases, the iterations progressively improve both training and testing error. Our algorithm terminates closer to $\ell=50$ in these cases.
The classical trade-off between over training and test error degradation is noticed in FICO and liver, and our algorithm terminates earlier than $\ell=50$ in these cases.




\bibliography{drlrules} 

\end{document}